\documentclass[conference, a4paper, top=0.75, bottom=1.0]{IEEEtran}
\IEEEoverridecommandlockouts
\usepackage{cite}
\usepackage{amsmath,amssymb,amsfonts}
\usepackage{algorithmic}
\usepackage{graphicx}
\usepackage{textcomp}
\usepackage{pifont}
\newcommand{\cmark}{\ding{51}}%
\newcommand{\xmark}{\ding{55}}%
\usepackage{xcolor}
\usepackage[cal=pxtx]{mathalfa}
\def\BibTeX{{\rm B\kern-.05em{\sc i\kern-.025em b}\kern-.08em
    T\kern-.1667em\lower.7ex\hbox{E}\kern-.125emX}}

\usepackage{fancyhdr}    
\usepackage{graphicx}
\usepackage{color}
\usepackage{comment}
\usepackage{subcaption}
\usepackage{multirow}
\usepackage{array}
\usepackage{hyperref}

\usepackage{floatrow}
\newcolumntype{P}[1]{>{\centering\arraybackslash}p{#1}}
\usepackage{framed}
\usepackage[normalem]{ulem}

\usepackage{xcolor}
\hypersetup{
    colorlinks,
    linkcolor={red!50!black},
    citecolor={blue!50!black},
    urlcolor={blue!80!black}
}

\begin{document}

\title{Learning To Segment Dominant Object Motion From Watching Videos
}


\author{\IEEEauthorblockN{Sahir Shrestha\IEEEauthorrefmark{1}\IEEEauthorrefmark{2}, Mohammad Ali Armin\IEEEauthorrefmark{2}, Hongdong Li\IEEEauthorrefmark{1} and
Nick Barnes\IEEEauthorrefmark{1}}
\IEEEauthorblockA{\IEEEauthorrefmark{1} The Australian National University\ \ \ \ \ \ \ \IEEEauthorrefmark{2} Data61, CSIRO\\
\texttt{\textit{firstname.lastname}}@\texttt{anu.edu.au},\ \ \ 
\texttt{ali.armin@data61.csiro.au}\\}
}

\def\Sahir#1{{\color{green}{\bf [Sahir:} {\it{#1}}{\bf ]}}}
\def\NB#1{{\color{red}{\bf [NB:} {\it{#1}}{\bf ]}}}
\def\Ali#1{{\color{blue}{\bf [Ali:} {\it{#1}}{\bf ]}}}

\maketitle

\begin{abstract}
Existing deep learning based unsupervised video object segmentation methods still rely on ground-truth segmentation masks to train. Unsupervised in this context only means that no annotated frames are used during inference. As obtaining ground-truth segmentation masks for real image scenes is a laborious task, we envision a simple framework for dominant moving object segmentation that neither requires annotated data to train nor relies on saliency priors or pre-trained optical flow maps. Inspired by a layered image representation \cite{layered1994}, we introduce a technique to group pixel regions according to their affine parametric motion. This enables our network to learn segmentation of the dominant foreground object using only RGB image pairs as input for both training and inference. We establish a baseline for this novel task using a new MovingCars dataset and show competitive performance against recent methods that require annotated masks to train. \footnote{Our code:
\href{https://github.com/sahir-shr/layersegnet}{\color{teal}\texttt{https://github.com/sahir-shr/layersegnet.}}}

\end{abstract}

\begin{IEEEkeywords}
deep learning, affine layered motion model, unsupervised video object segmentation
\end{IEEEkeywords}

\section{Introduction}

Motion cues provide us with a rich source of information for perception of our visual world. Objects and surfaces become immediately distinct once they start to move and we instinctively use motion information to group regions together \cite{koffka2013principles}. 
The task in Video Object Segmentation (VOS) is to separate the dominant foreground object(s) in a video sequence. Existing learning based approaches can be classified as either semi-supervised or unsupervised methods. 
In the semi-supervised \cite{lai2020mast, meinhardt2020make, perazzi2017learning, maninis2018video} setting, the ground-truth is provided for the first frame of the video during inference. The aim is then to propagate this mask and obtain segmentation for all subsequent frames. In unsupervised VOS \cite{mahadevan2020making, yang2019anchor, ren2021reciprocal}, no such object masks are available during inference. However, existing semi-supervised and unsupervised VOS methods still rely on ground-truth masks to train or use pre-trained optical flow \cite{ren2021reciprocal} or saliency models \cite{lu2020learning}.  In contrast, we propose a fully unsupervised training pipeline for our model that uses neither annotated labels nor pre-trained semantics as proxy supervision for training. Our network determines the dominant object in a scene through purely pixel motion analysis
 
A layered approach to representing 2D image sequences \cite{layered1994} provides a promising model for  
capturing the details of our 3D world where objects and surfaces move independently. This works by first estimating motions present, separating pixels into layers based on similarity of motion, and finally, moving layers on top of one another to synthesise the next image in the sequence. We adapt this process for a CNN by making the entire layered image synthesis pipeline differentiable for end-to-end training. We estimate two motions in a scene using affine parametric models and segment all image pixels into either of these motion classes. 


\begin{figure}[t!]
\centering
    \begin{subfigure}[b]{0.32\linewidth}
		\includegraphics[width=\textwidth]{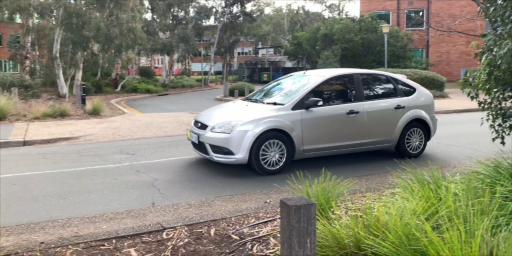}
	\end{subfigure}
	\begin{subfigure}[b]{0.32\linewidth}
		\includegraphics[width=\textwidth]{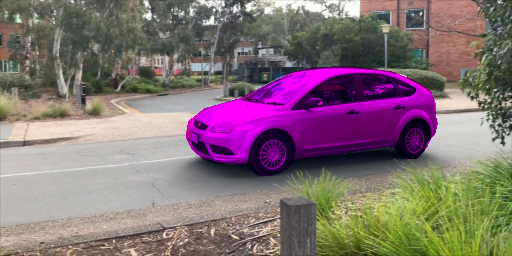}
	\end{subfigure}
	\begin{subfigure}[b]{0.32\linewidth}
		\includegraphics[width=\textwidth]{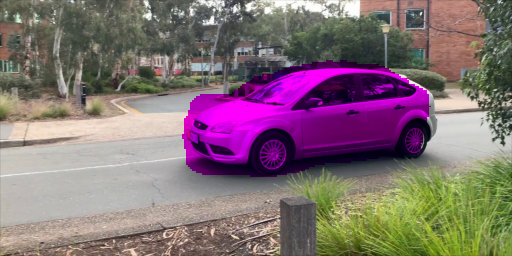}
	\end{subfigure}
	
    \begin{subfigure}[b]{0.32\linewidth}
		\includegraphics[width=\textwidth]{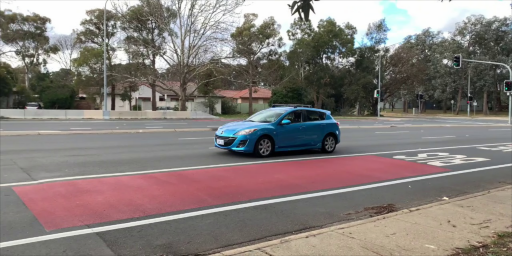}
		\subcaption{Frame1}
	\end{subfigure}
	\begin{subfigure}[b]{0.32\linewidth}
		\includegraphics[width=\textwidth]{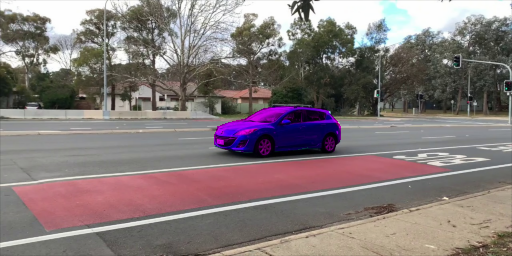}
				\subcaption{GT}
	\end{subfigure}
	\begin{subfigure}[b]{0.32\linewidth}
		\includegraphics[width=\textwidth]{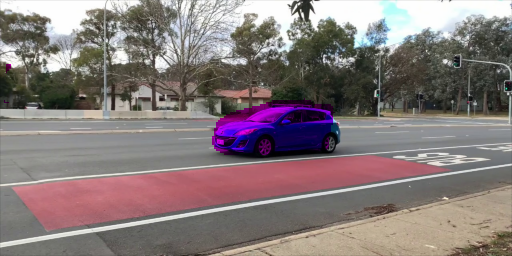}
				\subcaption{Prediction} 
	\end{subfigure}
	\caption{Our network successfully segments the dominant object in videos through a bottom-up grouping of pixels based on affine motion similarity. We train our network unsupervised using only unannotated RGB image pairs. Visual examples from two different scenes on MovingCars shown.}
	
\label{fig:intro}
\end{figure}

Image representation using moving layers is well-established in the literature \cite{darrell1991robust, jepson1993mixture, layered1994} but no implementation so far has employed an end-to-end CNN to achieve this. Sevilla-Lara \textit{et al.} in \cite{sevilla} combine a layered model and a neural network but their method is a hybrid approach where a pretrained network is combined with a variational expectation maximisation algrorithm for inference. In \cite{zhang2018layered}, Zhang \textit{et al.} use a maxout operation to perform disjoint separation of flow. 
However, they do not apply any explicit constraints during flow separation and it is unclear whether flow is reliably grouped based on motion homogeneity. In \cite{yang2021self}, Yang \textit{et al.} also propose a foreground object segmentation network that is trained without any manual annotations but they require a pre-computed flow map as input. They then use a slot attention network to group together pixels based on visual homogeneity. In contrast, our model does not need flow as input but only uses unannotated image pairs to group foreground/background pixels based on affine motion homogeneity. 


Using a layered image reconstruction process, we show that photometric loss can be used for learning to segment a dominant object in videos. We do not require pre-computed flow/saliency maps and use only unlabelled image sequences to train making our approach fully unsupervised during both training and inference.



We identify our three main contributions as follows:

\begin{enumerate}
    \item We propose a novel {\it unsupervised network
    to learn explicit dense dominant object segmentation} in an end-to-end CNN from only unlabelled pairs of RGB image sequences showing 2 rigid-body motions.
    \item We introduce a novel {\it layered differentiable image synthesis (LDIS)} module to separate frame 1 into two affine motion layers and synthesise frame 2. 
    \item We demonstrate that our network is able to explicitly segment the dominant foreground object without any annotation
    during training or inference. We contribute a new real-world dataset of moving cars and show competetive results against state-of-the-art methods. Our code and dataset will be made publicly available.  
\end{enumerate}

\section{Related Work}

\subsection{Video Object Segmentation}
 
VOS methods have traditionally relied on heuristics such as saliency \cite{wang2015saliency,guo2008spatio,mahadevan2009spatiotemporal},
object proposals using motion/appeareance information \cite{lee2011key, ma2012maximum,zhang2013video}
or motion analysis of point trajectories \cite{brox2010object,ochs2011object}.
With the advent of deep learning, the state-of-the-art has been dominated by semi-supervised methods that seek to propagate the given first frame object annotation across subsequent frames. Online methods perform network update during inference time to better capture object masks as the object moves through the sequence \cite{maninis2018video, meinhardt2020make}
whereas offline methods achieve this without requiring any network fine-tuning during inference \cite{li2018video,oh2018fast}.
In the unsupervised/self-supervised VOS setting, first frame object annotation is not provided and methods learn to detect and segment objects by exploiting appearance and spatio-temporal information \cite{tokmakov2019learning, jain2017fusionseg}.
Usually, pre-computed optical flow maps along with images are used as input in a two-stream framework \cite{ren2021reciprocal, zhou2020motion, fragkiadaki2015learning}
to generate object masks. Other methods use only image pairs as input and learn pairwise dependencies between frames using non-local operators to model temporal cues as done in AnchorDiff-VOS \cite{yang2019anchor}. Similarly, COSNet \cite{lu2019see} uses co-attention layers instead to capture rich correlations between frames. However, these methods still require ground-truth masks to train which can be laborious to annotate for real world images. 

In \cite{lu2020learning}, Lu \textit{et al.} train a VOS network without any annotated ground-truth data by relying on a saliency model and CAM maps to generate proxy-labels for foreground discrimination. However, their model only works in the semi-supervised setting. Yang \textit{et al.} in \cite{yang2021self} also train a VOS network without annotated labels but require a pre-computed flow map (that is converted to an RGB flow image) as input and use slot attention \cite{locatello2020object} to perform iterative clustering in RGB image space. 

In contrast, we train our network without any ground truth labels and require only RGB image sequences as input. Instead of relying on top-down saliency or object priors, we propose a fully unsupervised VOS framework that is based on a bottom-up approach to object segmentation by grouping together pixels based on affine motion similarity. 
 
\subsubsection{Layered Image Representation}
Methods incorporating layers for motion estimation and segmentation are well-established in literature with \cite{darrell1991robust, jepson1993mixture, layered1994} providing the first few seminal works. These methods exploit the idea that motion in any scene is not globally homogeneous but appears to be piecewise continuous. 
There have been many implementations that have since expanded on this idea \cite{ayer1995layered, hsu1994accurate, sun2013fully}.
Our main inspiration remains \cite{layered1994} where Wang and Adelson detail the process of layered image synthesis that we adapt for our end-to-end CNN implementation. This is in contrast to the expectation maximisation (EM) style algorithms that most methods use which can be tricky to optimise. Like \cite{jepson1993mixture, layered1994, lee1997layered}, we use an affine model to estimate motion for each layer. In this paper, we build a network to handle two motions.

There have been prior attempts to combine deep learning and a layered model with Sevilla-Lara \textit{et al.} in \cite{sevilla} using pre-trained scene semantics from a CNN to initialise layers. However, their method is not end-to-end or unsupervised and flow estimation is done using a variational EM algorithm (as in \cite{sun2013fully}). In contrast, our method uses a fully end-to-end CNN and only unlabelled RGB image sequences for training and inference. In \cite{zhang2018layered}, Zhang \textit{et al.} propose a CNN based layered optical flow estimation that relies on their ``soft-mask" module to separate of flow into disjoint classes but they do not synthesise images using layers. 
Our LDIS pipeline remains faithful to the traditional layered approach and provides explicit constraints during grouping of pixels using affine motion models allowing us to confidently identify motion homogeneous regions.


\subsubsection{Motion Segmentation}
Motion segmentation involves partitioning image pixels into groups with homogeneous motion. Typically, methods work towards separating pre-determined feature point trajectories that track moving objects through an image sequence \cite{brox2010object, arrigoni2019robust}.
It can be challenging to compute these tracks, requires an additional step, and the resulting segmentation is generally sparse. This problem has been tackled 
using geometric constraints \cite{jung2014rigid}, subspace clustering \cite{ji2015shape} and through fitting of motion models \cite{fischler1981random}. Another approach is to first compute optical flow to then cluster regions according to flow variations \cite{verri1989motion, fragkiadaki2015learning}. However, using flow as a precursor is likely to carry over errors intrinsic to the flow estimation process (e.g., inaccurate flow around motion boundaries due to smoothness regularisation). Some methods use pre-trained semantic information to initialise object regions \cite{zhuo2019unsupervised}. This approach 
relies on the accuracy of object recognition of the semantic network that is trained supervised using manual annotations. 
In \cite{ranjan2019competitive}, Ranjan \textit{et al.} obtain motion segmentation using an unsupervised CNN. However, they primarily exploit scene geometry to only distinguish between camera or independent motion, require camera parameters beforehand, and need five consecutive frames for inference. 

We propose an end-to-end CNN that can train without supervision or prior feature point trajectories - only two sequential RGB images are needed as input. We do not require a separate flow estimator to obtain segmentation - affine flow and segmentation are estimated in a symbiotic end-to-end framework. We take a bottom-up approach to motion segmentation by relying entirely on apparent pixel motion - no semantic priors or camera motion assumptions are necessary (as in \cite{ranjan2019competitive}). 

\section{Methodology}
In this section, we propose a video object segmentation framework that uses affine parametric models to distinguish the foreground from the background. We achieve this through a layered image representation whose theory is well-established in the literature but implementation in a deep neural network presents new challenges. A key challenge we tackle in this paper is ensuring that all operations within the architecture are differentiable for end-to-end learning.



\subsection{Layered Differentiable Image Synthesis} \label{sec:layered}
In \cite{layered1994}, Wang and Adelson describe a method to decompose an image sequence into layers that represent moving objects/surfaces. They propose using three maps to define each layer: the intensity (RGB) map, the alpha map (defines pixel opacity/transparency), and the velocity (optical flow) map. An image is decomposed into multiple disjoint layers using the intensity and alpha map. These layers are then warped (using the flow map) and reassembled to produce a reconstruction of the next image in the sequence.

\begin{figure}[t!]
\includegraphics[width=\textwidth]{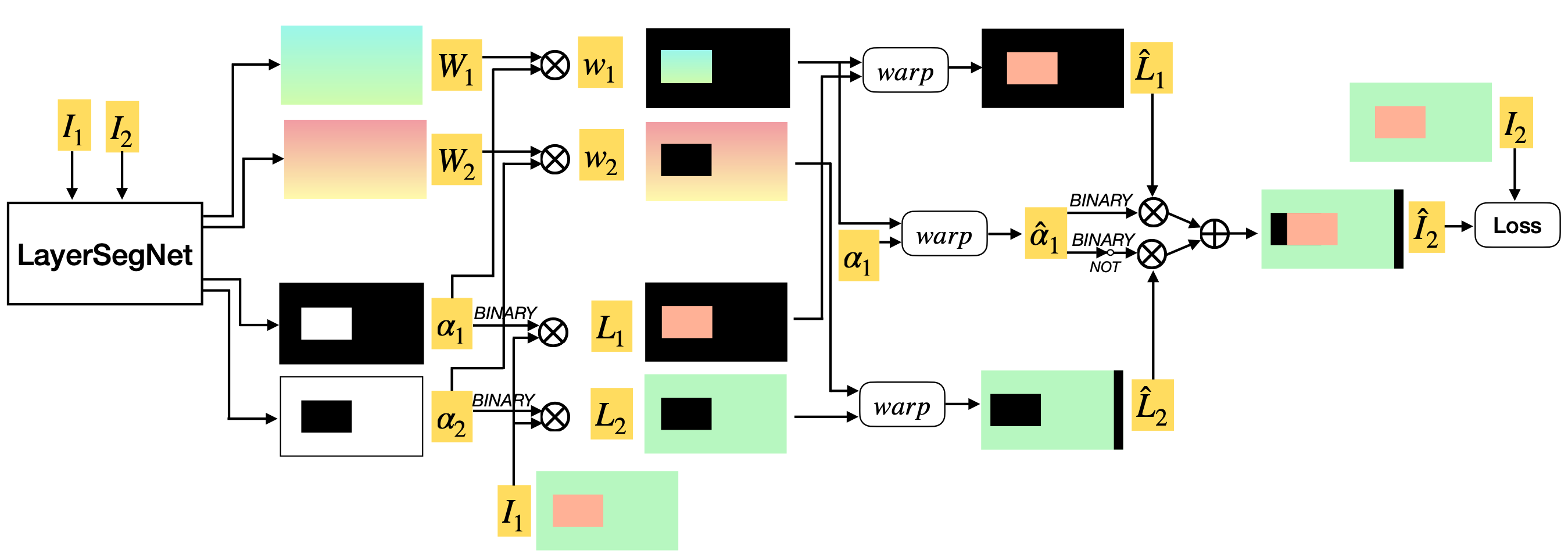}
\caption{Simulated visual example of our layered differentiable image synthesis pipeline where $I_1$ is separated into layers that are each individually warped
and combined to obtain $\hat{I}_2$.}
\label{fig:overall}
\end{figure}

Inspired by this process, we design a layered differentiable image synthesis (LDIS) module that we use to reconstruct frame 2 using frame 1 of an image pair. We assume that each pixel is either opaque or fully transparent and has an associated alpha value that provides pixel support (segmentation). Additionally, each layer also consists of an intensity map and a flow map that we use to composite layers for image synthesis. 
We show the overall image synthesis process in Figure \ref{fig:overall} and follow with details below.

Let us first define an input sequence of RGB images $I_1$ and $I_2$
with determinate and identical image size. We use the coordinate notation $\textbf{x} = (x,y)$ to access pixel values in the image domain of $I_1$ ($\Omega_1$) or $I_2$ ($\Omega_2$). Our aim is to separate $I_1$ into disjoint layers, move the layers independently and then combine the layers to synthesise a reconstruction of $I_2$. Each layer consists of an RGB intensity map ($L_i:  (\Omega \subset \mathbb{R}^2) \rightarrow \mathbb{R}^3$), an alpha map ($\alpha_i: (\Omega \subset \mathbb{R}^2) \rightarrow \mathbb{R}^1$) and a flow map ($w_i: (\Omega \subset \mathbb{R}^2) \rightarrow \mathbb{R}^2$). 

In the sequence $I_1, I_2$, let there be two distinct affine motions parameterised by $A_1$ and $A_2$:

\begin{align}
    W_i(x, y) = & 
    \textbf{A}_i
    \begin{bmatrix}
    1\\
    x\\
    y\\
    \end{bmatrix} =
    \begin{bmatrix}
    a_i^1 & a_i^2 & a_i^3\\
    a_i^4 & a_i^5 & a_i^6\\
    \end{bmatrix}
    \begin{bmatrix}
    1\\
    x\\
    y\\
    \end{bmatrix}
    ,\ i \in \{1, 2\} \label{eq:affine}
\end{align}
\noindent
where $W_i: (\Omega \subset \mathbb{R}^2) \rightarrow \mathbb{R}^2$ gives dense flow maps calculated using 
affine parameters $A_i$ for all \textbf{x}. Corresponding to the two affine motions, we separate all pixels in $I_1$ into layers using alpha maps that define spatial support regions. Ideally, for each pixel, we would have a one-hot vector and assign binary values
to indicate layer memberships (each pixel can only be associated with one motion). However, such discreet value assignment is not conductive to a differentiable framework. 

Instead, we implement softmax binning \cite{yang2018deep}
that approximates a hard binning operation.
Here, we use two `bins' for our two motion layers to obtain their alpha maps:

\begin{align}
    \alpha^i(\textbf{x}) \in [0,1], \ i \in \{1,2\} \label{eq:alphadefinition}
\end{align}
\noindent


We explicitly ensure that layer membership is disjoint using a modified maxout operation inspired by
\cite{zhang2018layered}. For each pixel, the maxout operation retains the maximal value of the two alpha maps, the non-maximal value is set to 0. We use these alpha maps to obtain the spatial support for flow of each layer:
\begin{align}
w_i = \alpha^i \odot W_i \label{eq:flowsupport}
\end{align}
where $\odot$ denotes element-wise product with broadcasting. Note that $W_i$ gives a dense calculation of flow values using affine parameters $A_i$ for all $\textbf{x}$ in the image domain.
The constrained layer flow map $w_i$ associates each pixel's alpha map to the motion of a particular layer. We also associate the alpha maps to pixel intensities.
However, to prevent input pixel intensities in $I_1$ being scaled by the continuous alpha map, we binarise these maps to separate $I_1$ into 2 layer intensity maps:
\begin{align}
L_i = \alpha^{i-binary} \odot I_1 \label{eq:layerintensity}\\ 
\alpha^{i-binary} = (\alpha^i > 0.5) \label{eq:alphabinary}
\end{align}
Note that binarisation of alpha map stops gradients from passing through this particular operation. However, 
the continuous alpha map in Eq. \ref{eq:flowsupport} allows
gradients
through during backpropagation.
This acts as a constraint on our network to effectively learn the alpha maps through association to the flow values in $W_i$ without affecting the pixel intensities in $I_1$. 

Next, we warp the intensity map ($L_i$) and alpha map ($\alpha_i$) of each layer using their flows $w_{i\in\{1, 2\}}$ with
forward warping \cite{niklaus2020softmax} to obtain the warped layer intensity maps $\hat{L}_i$:

\begin{align}
 \hat{L}_i(\textbf{x} + w_i(\textbf{x})) \leftarrow L_i(\textbf{x}) \label{eq:layerwarp}; \\
  \hat{\alpha}^i(\textbf{x} + w_i(\textbf{x})) \leftarrow \alpha^i(\textbf{x}); \label{eq:alphawarp} \\
  \forall \textbf{x} \in \textbf{m}_i,\ i \in \{1, 2\} \nonumber
\end{align}

We are ready to synthesise reconstruction of $I_2$ by compositing the warped layer intensity maps $\hat{L}_{i\in\{1,2\}}$. However, the warped layers will now contend for positions where they overlap. Our LDIS model explicitly resolves this through depth ordering via the alpha map
so that the layer closer to the camera occludes the layer beneath. This process mimics how moving objects result in occlusions.
This contention for output pixel position due to occlusion is an issue that 
arises in an explicit forward warping process.
In our case, we first perform warping separately for each layer then 
combine them using a fixed depth ordering. There is no issue of contention between pixels in the same layer due to our explicit affine motion model (i.e. no many-to-one mappings within a layer). Without a layered approach, a separate approach to occlusion reasoning is needed to solve this issue such as using brightness constancy as a measure of occlusion \cite{niklaus2020softmax}.
   
We synthesise $I_2$ from the warped layers by following a simple rule:
a smaller $i$ represents a layer closer to the camera (i.e. pixels of $\hat{L}_1$ occlude pixels of $\hat{L}_2$). 

Finally, we synthesise the output image $\hat{I}_2$. Since we only have 2 layers, with layer 1 on top of layer 2, we only need $\hat{\alpha}_1$ for reconstruction ($\hat{\alpha}_1$ gives the layer support for the warped pixels in $\hat{L}_1$). We binarise the warped alpha map as in Eq. \ref{eq:alphabinary} and reconstruct $I_2$ using the warped layers $\hat{L}_{i\in\{1, 2\}}$:

\begin{align}\label{eq:reconstruct}
    \hat{I}_2 =\ & \hat{\alpha}_{1-binary} \cdot \hat{L}_1\ + (1 - \hat{\alpha}_{1-binary}) \cdot \hat{L}_2 
\end{align}
\noindent
where $\hat{I}_2$ is the synthesised image. 

\subsection{Loss Function} \label{sec:loss}
When layers overlap, it is possible that for some pixel regions none of the layers contribute any pixel-value (resulting in black pixels in $\hat{I}_2$ in Figure \ref{fig:overall}). We detect these dis-occluded pixels by simply searching $\hat{I}_2$ for pixels that are zero-valued. The dis-occluded pixels are masked out during loss calculation using the dis-occlusion mask:

\begin{align}
\textbf{D}(\textbf{x}) = &
  \begin{cases}
    1 &,\ \text{if}\ \hat{I}_2(\textbf{x}) = 0 \\
    0 &,\ \text{otherwise}
  \end{cases}
\end{align}

The learning is driven by a photometric loss that 
uses the robust generalised Charbonnier penalty function \cite{bruhn2005lucas}: 


\begin{align}
Loss = & \sum_{\textbf{x} \in \Omega_2} (1 - \textbf{D}(\textbf{x})) \rho(I_2(\textbf{x}) - \hat{I_2}(\textbf{x})) \\
\rho(\textbf{x}) = & \sqrt{\textbf{x}^2 + 0.001^2} \label{eq-char}
\end{align}

We have now established the blueprint of our LDIS framework. We follow with specifics of implementation in a CNN.
  
\section{Network Implementation}\label{sec:implementation}
Our network LayerSegNet is inspired by encoder-decoder flow architectures \cite{dosovitskiy2015flownet,pwcnet}. We use the encoding structure from pwc-net \cite{pwcnet} that obtains separate feature encodings for $I_1$ and $I_2$. The outputs from the deepest level of the encoder are stacked then passed on to both the optical flow and segmentation pipelines. 
 
\begin{figure}[t!]
\includegraphics[width=\textwidth]{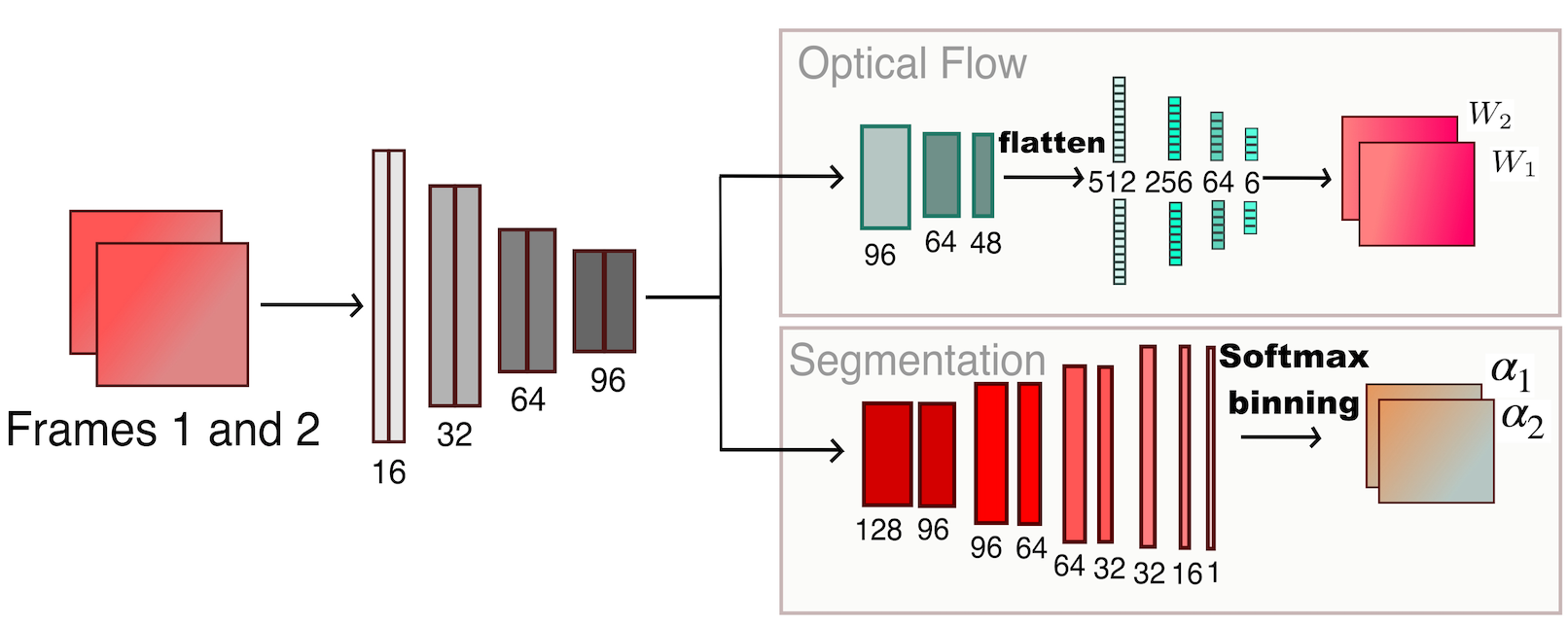}
\caption{Illustration of LayerSegNet that estimates flow maps ($W_1$, $W_2$) and alpha maps ($\alpha_1$, $\alpha_2$). Numbers beneath layers indicate number of feature channels.}\label{fig:network}
\end{figure}

\subsubsection{Optical Flow Module:}
It contains three convolutional layers (96-strided, 64-strided and 48 feature channels) followed by four fully connected layers (512, 256, 64 and six output units). Each convolution is followed by a leaky ReLU. This module outputs two sets of affine motion parameters: $\textbf{A}_1$ and $\textbf{A}_2$. Using these parameters and the process outlined in equation (\ref{eq:affine}) we obtain two flow maps $W_1,W_2 \in \mathbb{R}^{W\times H\times 2}$.

\subsubsection{Segmentation Module:}
This is a decoder module that takes the same input as the optical flow module and outputs a single value for each pixel $\textbf{x}$ that is continuous in the range $[0,1]$. It performs three bilinear upsamplings (followed by convolutions) to output a single-channeled map. 

This single-channeled map is passed through the softmax binning followed by maxout operation resulting in a dual-channeled map that at each pixel location gives the corresponding layer alpha value.

\begin{figure}[t!]
\centering
	\includegraphics[width=0.9\textwidth]{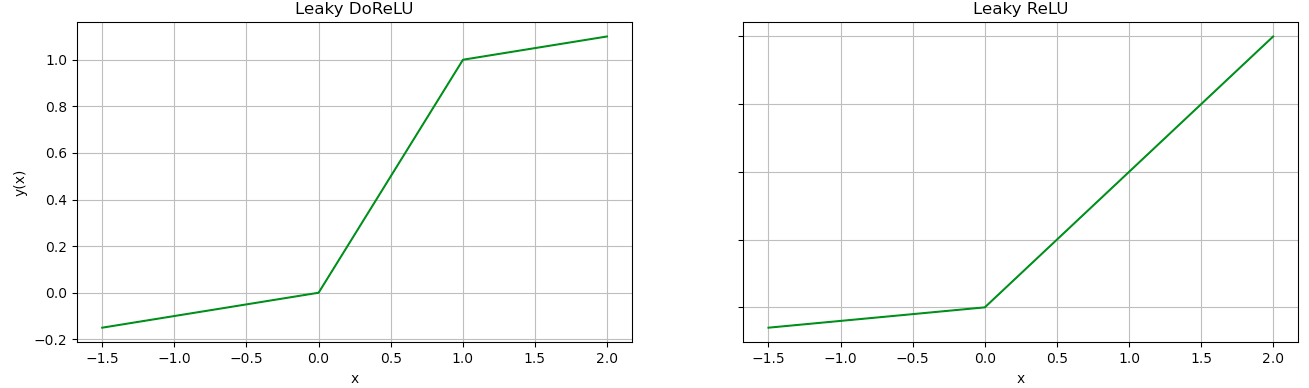}
\caption{Visualisation of our leaky DoReLU (left) vs leaky ReLU (right). The x-axis is input $x$ and y-axis is output $y(x)$.} \label{fig:drelu}
\end{figure}

In \cite{maas2013rectifier}, Maas {\it et al.} proposed the leaky ReLU that restricts the slope of the ReLU \cite{nair2010rectified} for input values below 0.
Empirically, we discovered that a modified leaky ReLU, when used as the activation function for estimating alpha maps, results in a significant boost to performance (see Section \ref{subsec:ablation}). We modify the leaky ReLU by restricting the output slope for inputs below 0 as well as above 1. We refer to it as the leaky double-rectified linear unit (leaky DoReLU), show a comparison between leaky ReLU and leaky DoReLU in Figure \ref{fig:drelu}, and define it as follows:

\begin{align}
    y(x) = &
  \begin{cases}
    1 + \frac{x - 1}{\gamma} &,\ x > 1 \\
    x &,\ \text{if}\ 0 \leq x \leq 1 \\
    \frac{x}{\gamma} &,\ x < 0.
  \end{cases}
\end{align}

Note that our leaky DoReLU imposes leaky upperbound and lowerbound restrictions to the input signal and is different in implementation compared to the dual rectified linear unit, \cite{godin2018dual}.
In this module, each convolution is followed by a leaky DoReLU with $\gamma = 10$ except the last activation function that caps values to be strictly within $[0,1]$ (leaky part is removed) giving the required input to the softmax binning operation.

\subsubsection{Image Reconstruction:}
Using $W_1$, $W_2$, $\alpha_1$, $\alpha_2$ and input image $I_1$, we perform the following steps to reconstruct $I_2$:


\begin{enumerate}
\item Extract the flow for each layer (eq.~\ref{eq:flowsupport}): $w_{i} = \alpha^i * W_i$
\item Extract layer intensity maps (eq.~\ref{eq:layerintensity}): $L_i = \alpha^{i-binary} * I_1; \alpha^{i-binary} = (\alpha^i > 0)$
\item Warp layer intensity maps with flows (eq.~\ref{eq:layerwarp}): $\hat{L}_i = \textbf{warp}(L_i, w_{i})$
\item Warp $\alpha^1$ with flow $w_{1}$ (eq.~\ref{eq:alphawarp}): $\hat{\alpha}^1 = \textbf{warp}(\alpha^1, w_{1})$
\item Combine warped layers (eq.~\ref{eq:reconstruct}): $\hat{I}_2 = \hat{\alpha}^{1-binary} \cdot \hat{L}_1\ + (1 - \hat{\alpha}^{1-binary}) \cdot \hat{L}_2$
\end{enumerate}
\noindent
where $i=\{1,2\}$ is the layer index, $\alpha^{i-binary}$ gives a binarised alpha map and \textbf{warp()} is  the forward warping process.

\section{Experiments}

\begin{figure*}[t!]
\centering
 
	\begin{subfigure}[b]{0.19\linewidth}
		\includegraphics[width=\textwidth]{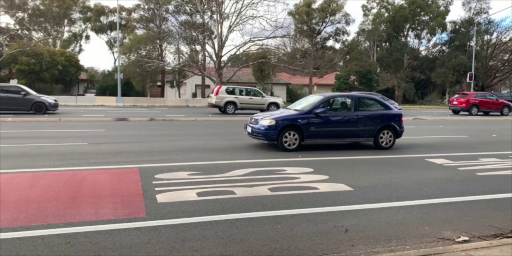}
	\end{subfigure}
	\begin{subfigure}[b]{0.19\linewidth}
		\includegraphics[width=\textwidth]{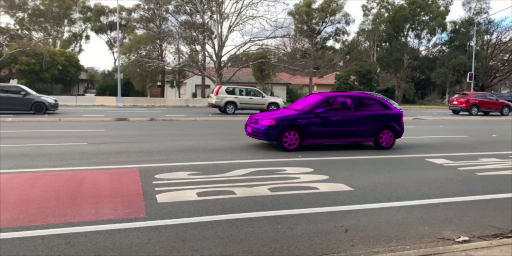}
	\end{subfigure}
	\begin{subfigure}[b]{0.19\linewidth}
		\includegraphics[width=\textwidth]{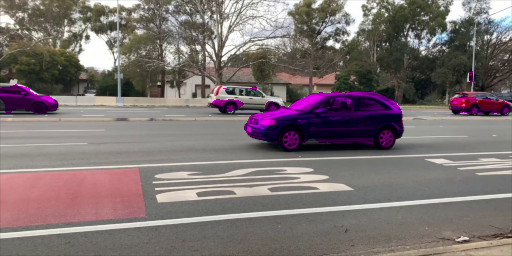}
	\end{subfigure}
	\begin{subfigure}[b]{0.19\linewidth}
		\includegraphics[width=\textwidth]{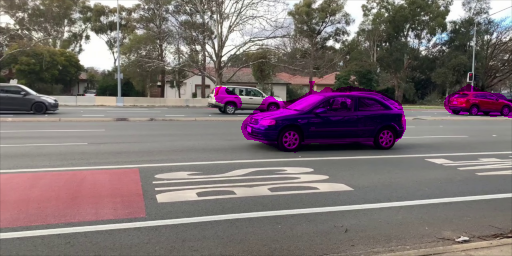}
	\end{subfigure}
	\begin{subfigure}[b]{0.19\linewidth}
		\includegraphics[width=\textwidth]{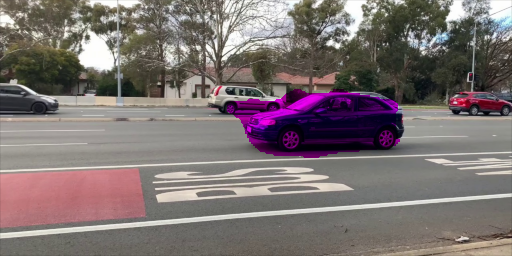}
	\end{subfigure}
	  
	\begin{subfigure}[b]{0.19\linewidth}
		\includegraphics[width=\textwidth]{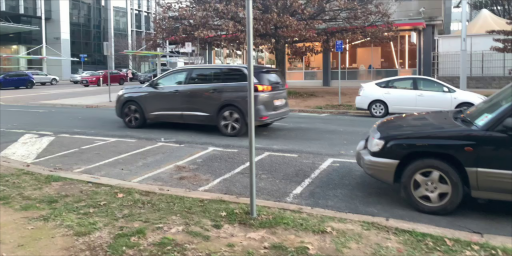}
	\end{subfigure}
	\begin{subfigure}[b]{0.19\linewidth}
		\includegraphics[width=\textwidth]{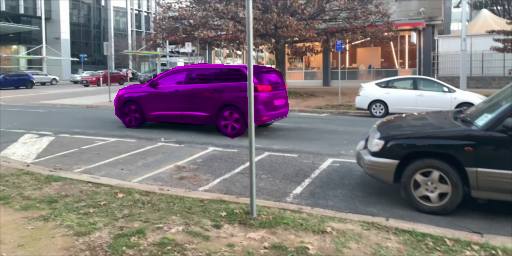}
	\end{subfigure}
	\begin{subfigure}[b]{0.19\linewidth}
		\includegraphics[width=\textwidth]{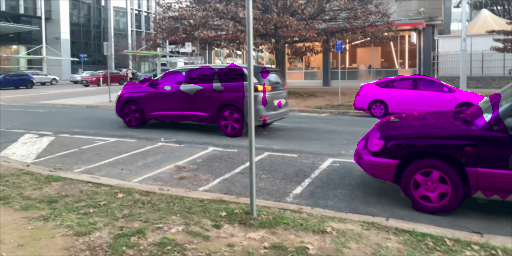}
	\end{subfigure}
	\begin{subfigure}[b]{0.19\linewidth}
		\includegraphics[width=\textwidth]{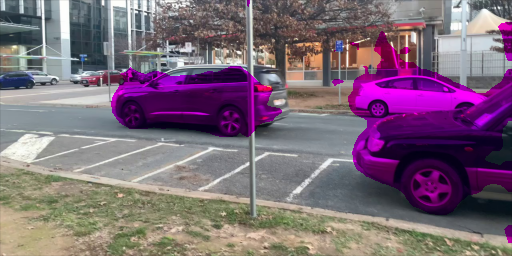}
	\end{subfigure}
	\begin{subfigure}[b]{0.19\linewidth}
		\includegraphics[width=\textwidth]{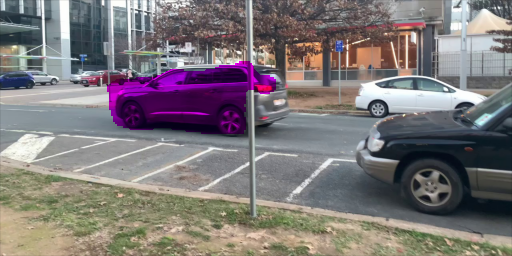}
	\end{subfigure}
	 
	\begin{subfigure}[b]{0.19\linewidth}
		\includegraphics[width=\textwidth]{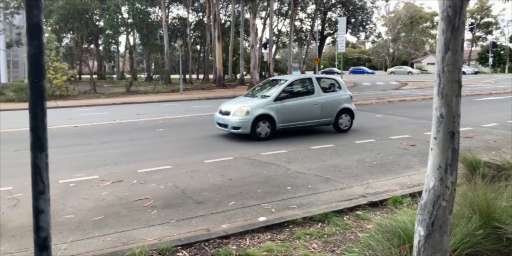}
	\end{subfigure}
	\begin{subfigure}[b]{0.19\linewidth}
		\includegraphics[width=\textwidth]{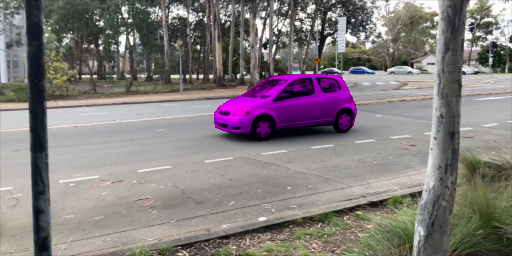}
	\end{subfigure}
	\begin{subfigure}[b]{0.19\linewidth}
		\includegraphics[width=\textwidth]{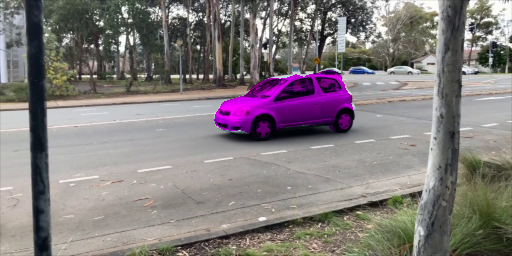}
	\end{subfigure}
	\begin{subfigure}[b]{0.19\linewidth}
		\includegraphics[width=\textwidth]{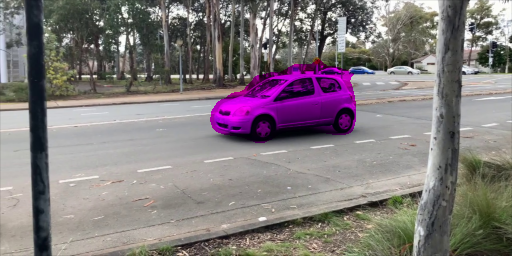}
	\end{subfigure}
	\begin{subfigure}[b]{0.19\linewidth}
		\includegraphics[width=\textwidth]{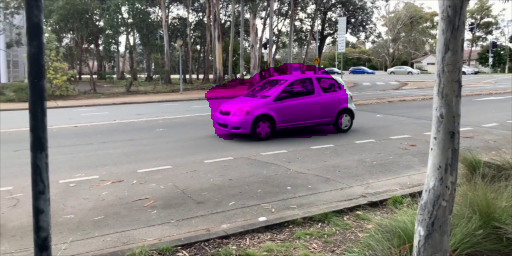}
	\end{subfigure}
	   
	\begin{subfigure}[b]{0.19\linewidth}
		\includegraphics[width=\textwidth]{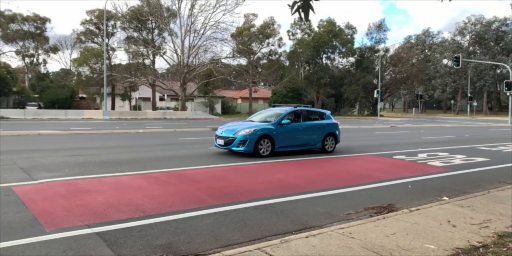}
	\end{subfigure}
	\begin{subfigure}[b]{0.19\linewidth}
		\includegraphics[width=\textwidth]{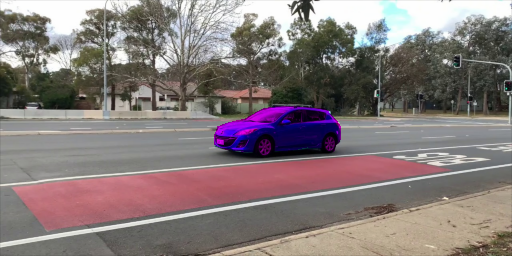}
	\end{subfigure}
	\begin{subfigure}[b]{0.19\linewidth}
		\includegraphics[width=\textwidth]{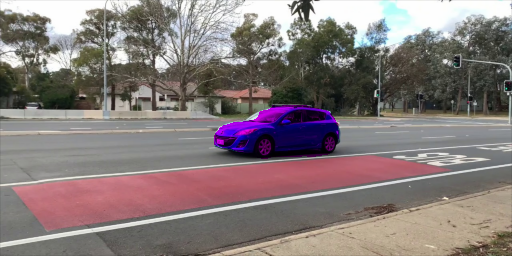}
	\end{subfigure}
	\begin{subfigure}[b]{0.19\linewidth}
		\includegraphics[width=\textwidth]{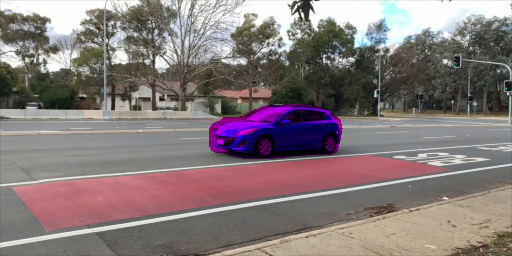}
	\end{subfigure}
	\begin{subfigure}[b]{0.19\linewidth}
		\includegraphics[width=\textwidth]{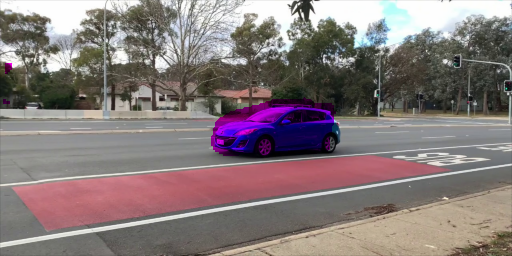}
	\end{subfigure}
	  
    \begin{subfigure}[b]{0.19\linewidth}
		\includegraphics[width=\textwidth]{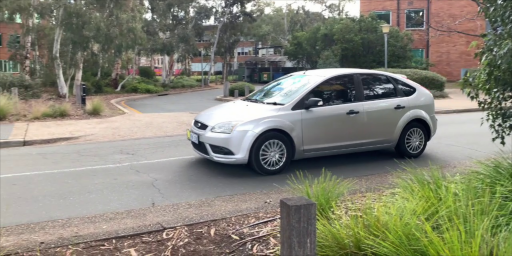}
		\subcaption*{Frame1}
	\end{subfigure}
	\begin{subfigure}[b]{0.19\linewidth}
		\includegraphics[width=\textwidth]{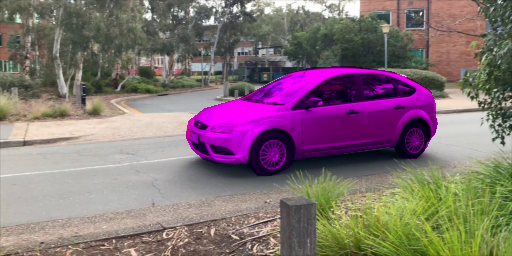}
		\subcaption*{GT}
	\end{subfigure}
	\begin{subfigure}[b]{0.19\linewidth}
		\includegraphics[width=\textwidth]{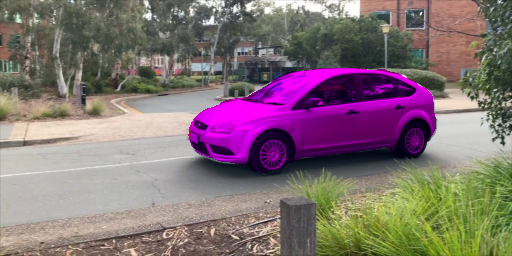}
		\subcaption*{AnchorDiff-VOS}
	\end{subfigure}
	\begin{subfigure}[b]{0.19\linewidth}
		\includegraphics[width=\textwidth]{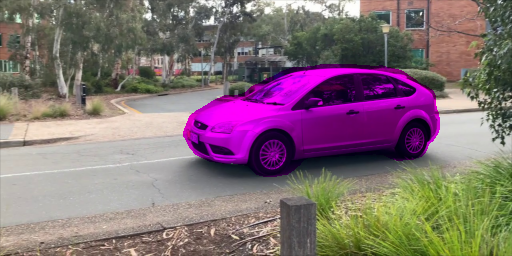}
		\subcaption*{COSNet}
	\end{subfigure}
	\begin{subfigure}[b]{0.19\linewidth}
		\includegraphics[width=\textwidth]{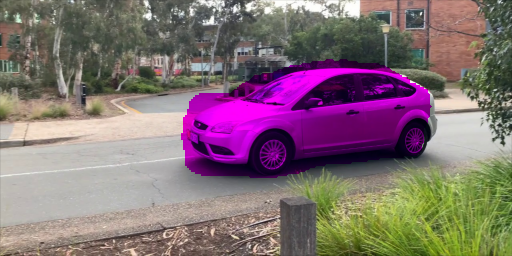}
		\subcaption*{LayerSegNet (Ours)}
	\end{subfigure}
\caption{Results on the MovingCars dataset of LayerSegNet (ours) trained without any annotated masks purely on a synthetic dataset (no finetuning) against AnchorDiff-VOS \cite{yang2019anchor} and COSNet \cite{lu2019see}. LayerSegNet, due to its purely affine motion segmentation approach successfully segments the dominant moving car in Rows 1 and 2 where AnchorDiff-VOS and COSNet both label all instances of cars (even cars that are not moving as in Row 2) due to their reliance on object features.}
\label{fig:results_mov_cars}
\end{figure*}
 
\subsection{Datasets}


We train our network on a synthetic dataset named RoamingSeg, created using images from \cite{janai}, where a randomly scaled foreground moves on top of a moving background (all movements are random and linear). We create 11,000 sequences of resolution 256x512 (HxW) with ground-truth flow and segmentation maps. 
 
We create a testing dataset named MovingCars to demonstrate object segmentation on real-world car images with affine motions. In each video scene, a dominant foreground object (car) is moving across the background. We prepare five sequences of moving cars with different backgrounds. The videos were taken handheld using a mobile phone camera that pans along with the moving cars.  We do not finetune our network on MovingCars to test the generalisation capability of our method on real world street scenes.


We also create another dataset named MovingObjects under more controlled setting to demonstrate dense motion segmentation on real-world images with two rigid-body motions. In the image scene, a foreground object and a background surface both undergo small translations. We prepare a fine-tuning set for training that contains sequences of three objects 
against a common background. We build a test set that adds four additional unseen object classes set against unseen backgrounds. The images were taken using a mobile phone camera that was rested on and moved along a solid surface. For both these datasets, each class in the test set contains three sequential images with ground-truth segmentation masks for the first two images. 




 \subsection{Training Details}
We use a batchsize of 8 and train our network end-to-end using adam \cite{kingma2014adam} with $\beta_1=0.9$ and $\beta_2=0.999$. Similar to pwc-net, our network outputs a quarter-resolution flow and segmentation maps that are upsampled to input image size to improve training efficiency. On RoamingSeg, we train for 600,000 iterations and start with a learning rate of $1.25\times 10^{-5}$. We use this network to test on the MovingCars dataset (without any finetuning) comparing against two recent VOS methods that are trained with annotated ground-truth masks. We found no methods that directly match our approach of VOS using using unlabelled image pairs during both training and inference.  


To finetune on MovingObjects, we train for 400,000 iterations with a learning rate of $6.25\times 10^{-6}$.
The learning rates are halved every 200,000 iterations. Note that we train our network from scratch fully unsupervised using our photometric loss. 
At no stage of training do we make use of any semantic information or explicit supervision using ground-truth maps. 

\subsection{Evaluation}
We evaluate our network using the mean Jaccard Index $\mathcal{J}$ which is the intersection over union of predicted and GT object masks.
We demonstrate our network's ability to segment the dominant foreground object
on MovingCars (Figure \ref{fig:results_mov_cars}), show results for rigid-body motion segmentation on MovingObjects (Figure \ref{fig:results}) and perform an ablation study for our DoReLU activation (Table \ref{table:ablation}). 


\subsection{LayerSegNet vs Maxout baseline}
To the best of our knowledge, our method is the first implementation of an end-to-end layered image model in a CNN. 
In \cite{zhang2018layered}, Zhang \textit{et al.}
introduce their ``softmask" module that performs layered flow estimation using a CNN but without layered image modelling. This was achieved through a modified maxout operation 
to obtain $k$ disjoint soft-masks that are multiplied to $k$ estimated flow maps ($k$ determined beforehand). These resulting disjoint flow maps are added to obtain the final flow. The authors state that the softmask module can be used with any base flow estimation network to improve flow.

We use the same modified maxout operation in our network to ensure that the estimated alpha maps are disjoint. However, the ability of our network to achieve motion segmentation comes from our overall LDIS module. We posit that obtaining disjoint flow maps using maxout is not sufficient to obtain segmentation of motion-homogeneous regions. To investigate this and compare motion segmentation results against our LDIS framework, we create a baseline network using the softmask module.



Since we were unable to obtain source code for \cite{zhang2018layered}, we reproduce a similar network by modifying our network to accomodate the softmask module while removing LDIS. The segmentation module is changed to output two soft-masks and the optical flow module to give two dense flow maps (as in \cite{zhang2018layered}). The soft-masks are made disjoint using maxout then multiplied to the flow maps and fused to obtain a final (single) flow map. We train this network supervised on RoamingSeg using ground-truth flow and on the same training regimen as our LayerSegNet. 

\begin{figure}[t!]
\centering
    \begin{subfigure}[b]{0.19\linewidth}
		\includegraphics[width=\textwidth]{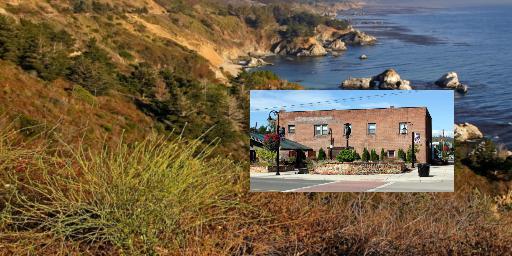}
	\end{subfigure}
    \begin{subfigure}[b]{0.19\linewidth}
		\includegraphics[width=\textwidth]{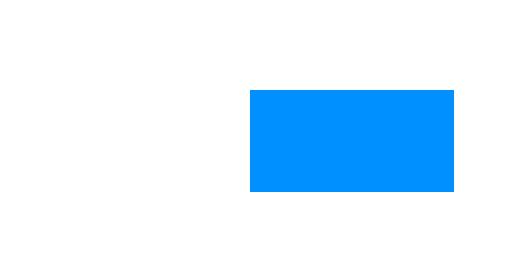}
	\end{subfigure}
    \begin{subfigure}[b]{0.19\linewidth}
		\includegraphics[width=\textwidth]{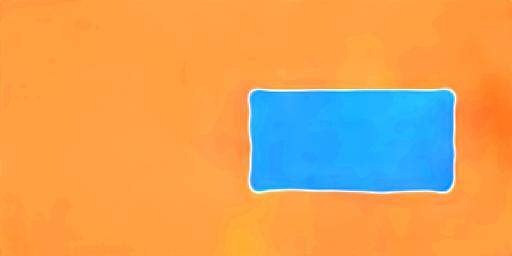}
	\end{subfigure}
    \begin{subfigure}[b]{0.19\linewidth}
		\includegraphics[width=\textwidth]{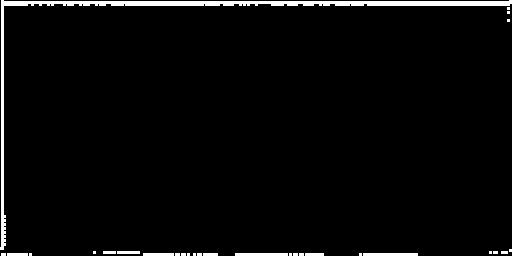}
	\end{subfigure}
    \begin{subfigure}[b]{0.19\linewidth}
		\includegraphics[width=\textwidth]{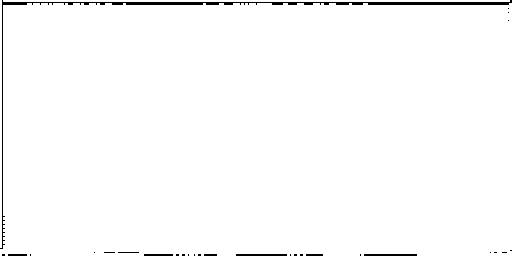}
	\end{subfigure}
	
	\begin{subfigure}[b]{0.19\linewidth}
		\includegraphics[width=\textwidth]{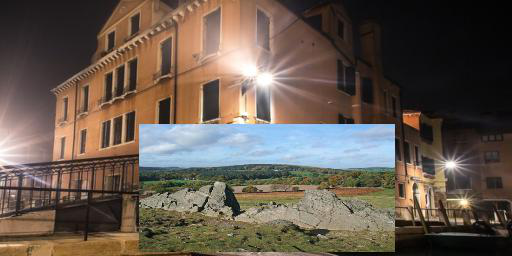}
		\subcaption{}
	\end{subfigure}
    \begin{subfigure}[b]{0.19\linewidth}
		\includegraphics[width=\textwidth]{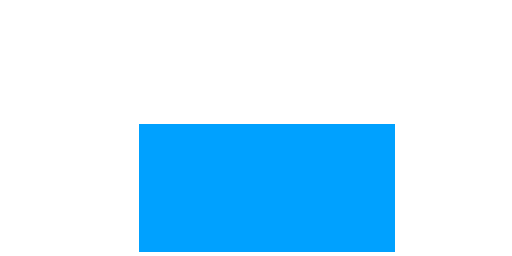}
		\subcaption{}
	\end{subfigure}
    \begin{subfigure}[b]{0.19\linewidth}
		\includegraphics[width=\textwidth]{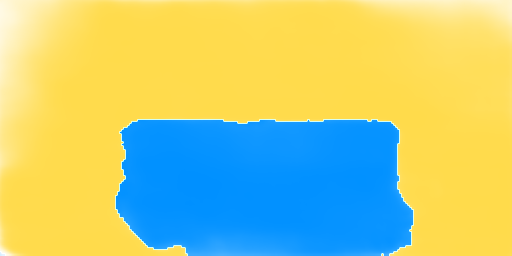}
		\subcaption{}
	\end{subfigure}
    \begin{subfigure}[b]{0.19\linewidth}
		\includegraphics[width=\textwidth]{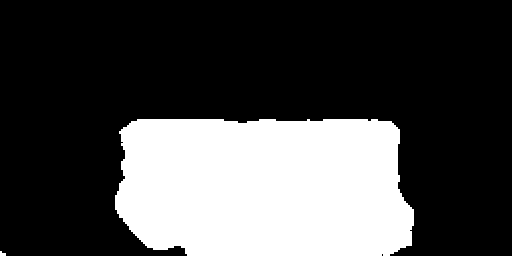}
		\subcaption{}
	\end{subfigure}
    \begin{subfigure}[b]{0.19\linewidth}
		\includegraphics[width=\textwidth]{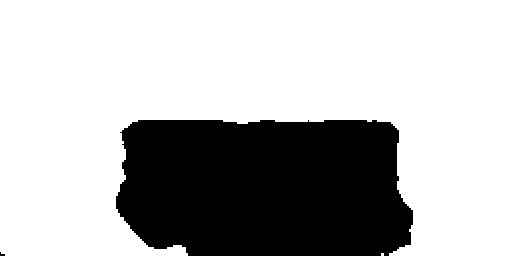}
		\subcaption{}
	\end{subfigure}
	
\caption{Visual comparison of results between maxout baseline (top row) and our LayerSetNet (bottom row) on RoamingSeg. From the left column: (a) Frame 1, (b) Ground-truth flow, (c) Estimated flow, (d) Layer 1, (e) Layer 2.}
\label{fig:ablation}
\end{figure}


In the maxout baseline, we find that although the flow is accurately estimated, there is no visible separation of motion observable in the softmasks (Figure \ref{fig:ablation}). We observe that the network only uses one intermediate flow map to estimate the final flow as evident by the lack of activations in Layer 1 and complete activations in Layer 2. In contrast, our proposed network successfully separates the foreground from the background (bottom row in Figure \ref{fig:ablation}). 


\subsection{Results}
\subsubsection{MovingCars}
On the MovingCars dataset, we compare our network LayerSegNet against two recent unsupervised VOS methods that are most similar to our approach: they are both end-to-end and do not need any annotations during inference. However, unlike us, they both require annotated masks during training. 
AnchorDiff-VOS \cite{yang2019anchor} performs feature propagation across image sequences using non-local operators while COSNet \cite{lu2019see} uses co-attention layers for global matching correspondences across images.
We show qualitative results in Figure \ref{fig:results_mov_cars} where we show competitive performance against these methods. Our LayerSegNet performs purely bottom-up segmentation of dominant object based on affine motions whereas AnchorDiff-VOS and COSNet are both more reliant on feature similarity. We postulate this is why, in Row 2, we find that both of these methods have erroneously segmented the two parked cars on the right half of the image. Those cars do not move and only the car at the center (as seen in GT) shows motion. Our network does not give this false positive result and only segments the moving car.
\begin{table}[t]
\centering
\begin{tabular}{P{3cm} P{2cm} P{2cm}}
Method & Supervised Training & MovingCars ($\mathcal{J}$ Mean $\uparrow$)\\ 
\hline\hline
COSNet \cite{lu2019see} & \cmark & 0.68 \\
AnchorDiff-VOS\cite{yang2019anchor} & \cmark & 0.73\\
\hline
LayerSegNet \textbf{Ours} & \xmark & 0.70\\
\hline
\end{tabular}
\caption{Quantitative results of our Network LayerSegNet compared against two recent VOS methods on MovingCars.}\label{table:main}
\end{table}

Similarly, in Row 1, we find that both AnchorDiff-VOS and COSNet attempt to segment all cars in the image whereas ours can identify the dominant foreground object in the scene. In other scenes (Rows 3-5) where only a single moving car is present, we show comparable results to these two methods. Quantitative results on MovingCars in Table \ref{table:main} show that our network outperforms COSNet by 2.9\% and is only 4.2\% behind AnchorDiff-VOS. Our LayerSegNet uses no annotated masks to train and is trained purely on a synthetic dataset with zero finetuning. In contrast, both AnchorDiff-VOS and COSNet rely on extensive ground-truth segmentation masks to train their networks. Note that our approach requires the dominant object to be moving with affine motion. If a car is moving more directly toward the camera its motion will not be affine, and the vehicle may only be partially detected. Examples of failed scenes are shown in the supplementary material.

\begin{figure}[t!]
\centering
    \begin{subfigure}[b]{0.3\linewidth}
		\includegraphics[width=\textwidth]{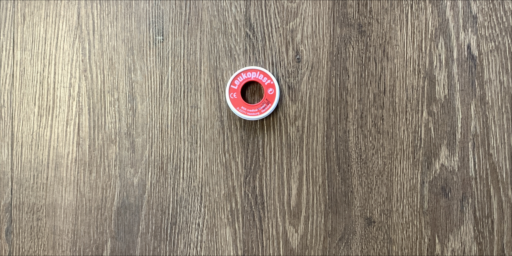}
	\end{subfigure}
	\begin{subfigure}[b]{0.3\linewidth}
		\includegraphics[width=\textwidth]{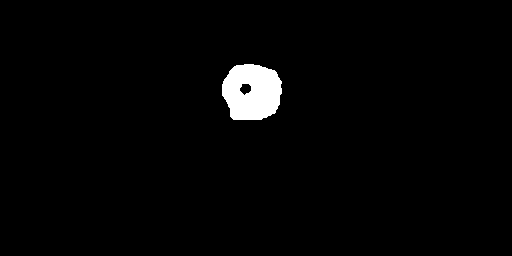}
	\end{subfigure}
	\begin{subfigure}[b]{0.3\linewidth}
		\includegraphics[width=\textwidth]{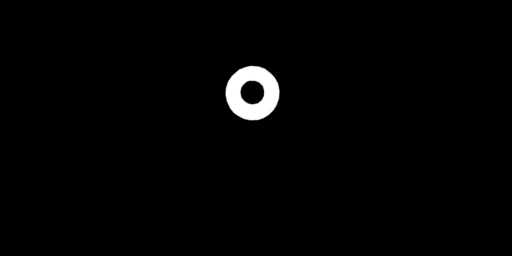}
	\end{subfigure}
	
	\begin{subfigure}[b]{0.3\linewidth}
		\includegraphics[width=\textwidth]{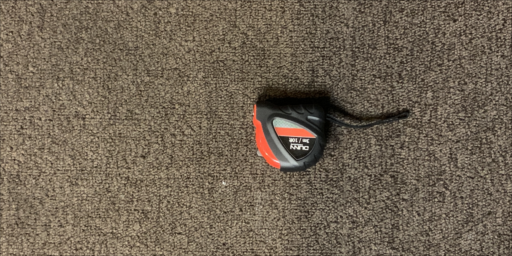}
	\end{subfigure}
	\begin{subfigure}[b]{0.3\linewidth}
		\includegraphics[width=\textwidth]{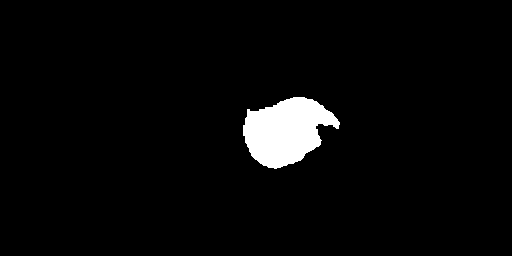}
	\end{subfigure}
	\begin{subfigure}[b]{0.3\linewidth}
		\includegraphics[width=\textwidth]{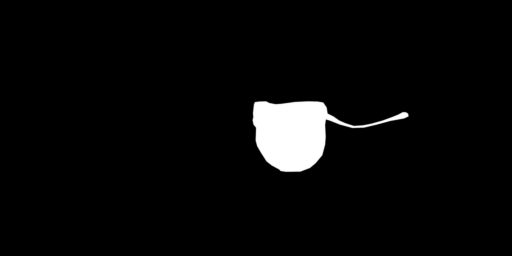}
	\end{subfigure}



	\begin{subfigure}[b]{0.3\linewidth}
		\includegraphics[width=\textwidth]{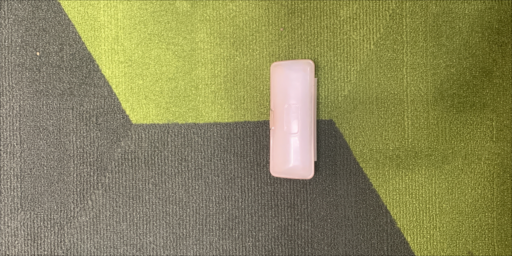}
	\end{subfigure}
	\begin{subfigure}[b]{0.3\linewidth}
		\includegraphics[width=\textwidth]{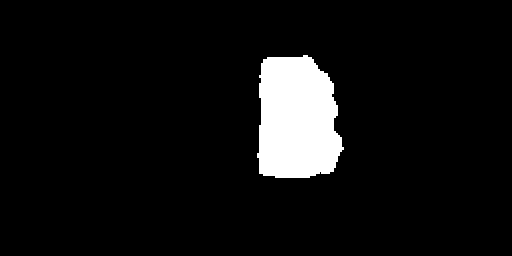}
	\end{subfigure}
	\begin{subfigure}[b]{0.3\linewidth}
		\includegraphics[width=\textwidth]{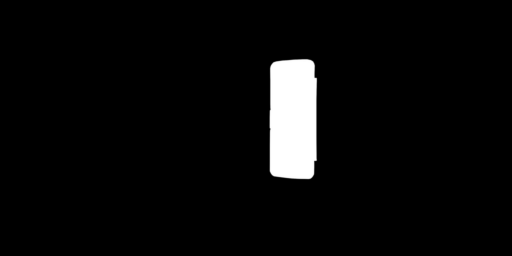}
	\end{subfigure}


	\begin{subfigure}[b]{0.3\linewidth}
		\includegraphics[width=\textwidth]{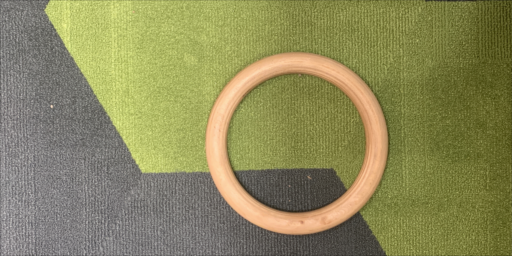}
	\end{subfigure}
	\begin{subfigure}[b]{0.3\linewidth}
		\includegraphics[width=\textwidth]{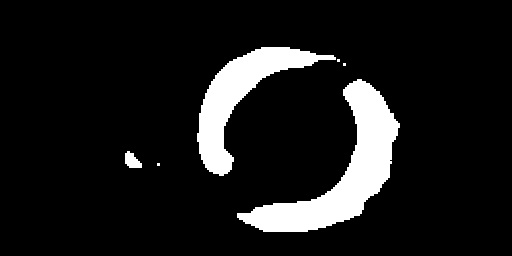}
	\end{subfigure}
	\begin{subfigure}[b]{0.3\linewidth}
		\includegraphics[width=\textwidth]{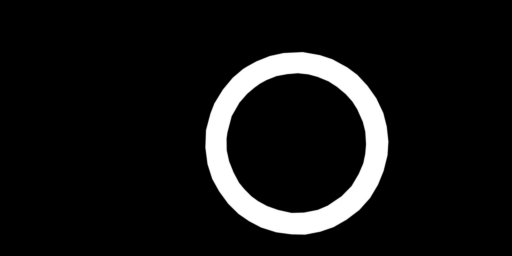}
	\end{subfigure}

	\begin{subfigure}[b]{0.3\linewidth}
		\includegraphics[width=\textwidth]{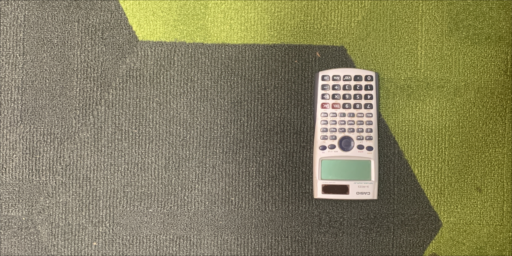}
	\end{subfigure}
	\begin{subfigure}[b]{0.3\linewidth}
		\includegraphics[width=\textwidth]{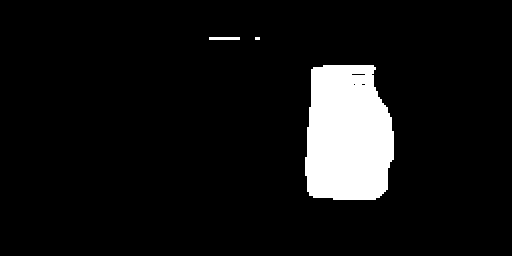}
	\end{subfigure}
	\begin{subfigure}[b]{0.3\linewidth}
		\includegraphics[width=\textwidth]{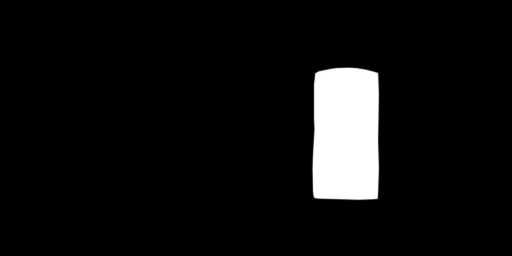}
	\end{subfigure}
	
	 
\caption{Our results on some classes of MovingObjects show good segmentation of rigid-body motions for a variety of objects (Rows 1-2 are objects not seen during finetuning).}
\label{fig:results}
\end{figure}

\subsubsection{MovingObjects}
We show quantitative results for MovingObjects in Table \ref{table:ablation}. Leaky DoReLU (ours) shows strong performance in terms of Jaccard mean. Visual inspection of the segmentation results in Figure \ref{fig:results} shows favourable results. Moving objects are correctly identified including for unseen objects (rows 1-4) and unseen backgrounds (rows 1-3). The segmentation maps show approximate coverage of the rectangular object classes. The gymnastic ring (fifth row) shows incomplete segmentation. As it has little internal texture, it may be that the motion is not consistently recovered.
 The maxout baseline failed to produce any foreground segmentation. 

\subsection{Ablation}

\begin{figure}[t]
\centering
    \begin{subfigure}[b]{0.29\linewidth}
		\includegraphics[width=\textwidth]{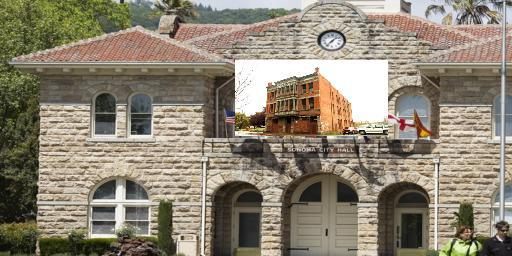}
	\end{subfigure}
    \begin{subfigure}[b]{0.29\linewidth}
		\includegraphics[width=\textwidth]{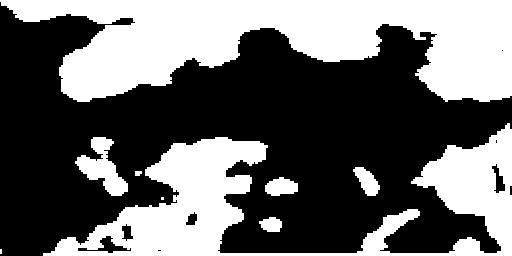}
	\end{subfigure}
    \begin{subfigure}[b]{0.29\linewidth}
		\includegraphics[width=\textwidth]{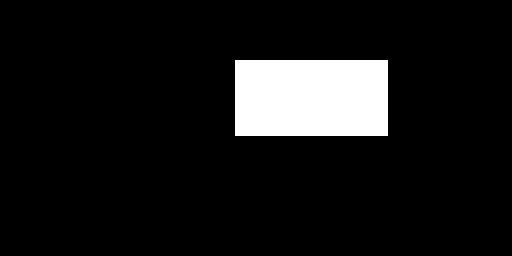}
	\end{subfigure}
	 
\caption{Visual example of a failed segmentation due to scaling of $I_1$ pixel intensities by continuous-valued $\alpha$ maps. From the left column: (a) Frame 1, (b) Estimated segmentation, (c) GT.}
\label{fig:failures}
\end{figure}

Our network fails to learn to segment motions without many vital components of our approach.
For example, we investigated relaxing the affine model to a regular optical flow map with smoothness regularisation for each layer. However, the network failed to learn accurately separate pixels into motion layers - we suspect applying only smoothness penalty is insufficient constraint towards segmenting pixels into motion layers.
When separating $I_1$ pixels into layers, we found that binarising the $\alpha$ maps is crucial as this prevents the pixel intensities from being scaled by the continuous $\alpha$ values. This ensures that Frame 2 can be recreated without the network having to learn to modulate the pixel intensities. Without this change, the network fails to achieve any segmentation at all (Figure \ref{fig:failures}).

\subsubsection{Leaky ReLU vs Leaky DoReLU activation}\label{subsec:ablation}
As reported in Section \ref{sec:implementation}, we use a leaky double rectified linear unit (DoReLU) as activation in our segmentation module. Two LayerSegNets, identical except for the activation (leaky DoReLU vs leaky ReLU) in the segmentation module, were trained and their results compared. 

On MovingObjects, we obtained a mean IoU of 0.67 for leaky DoReLU (ours) vs 0.40 for leaky ReLU - a performance increase of 67.5\%. Since our alpha map values are enforced to $[0,1]$, we postulate that the reason for the improved performance of leaky DoReLU is due to stable gradients when activation outputs are restricted below 0 and above 1 (compared to leaky ReLU that has no upperbound restriction).

\begin{table}[h]
\centering
\begin{tabular}{P{3.5cm} P{2.5cm} P{2cm}}
Activation & $\mathcal{J}$ Mean $\uparrow$ & $\Delta \mathcal{J}$(\%)\\
\hline\hline
Leaky ReLU & 0.40 & 0.0\%\\
Leaky DoReLU (ours) & 0.67 & +67.5\%\\
\hline
\end{tabular}
\caption{Comparing segmentation accuracy of our LayerSegNet using leaky ReLU vs leaky DoReLU activation on the MovingObjects test set.
}\label{table:ablation}
\end{table}

\section{Conclusions}
We propose a bottom-up approach to segmenting a dominant foreground object in videos by grouping pixels based on affine motion. By leveraging a representation of images using moving layers, we successfully obtain dense segmentation without requiring pre-trained saliency priors, optical flow maps or any manual annotations during training/inference.

Our two layered affine motion model works remarkably well on the real world MovingCars dataset where we show favourable results against methods that were trained using annotated ground-truths. 
We hope our demonstration on our datasets will prove beneficial by drawing attention and enabling further innovations to this novel task of segmenting dominant moving object in videos without any human supervision. 
To this end, we make our model and datasets publicly available.



\clearpage




\section*{Supplementary Material}

\subsection{Forward Warping Formulation}\label{sec:fwdwarp}

In \cite{jaderberg}, Jaderberg \textit{et al.} propose a spatial transformer module that facilitates differentiable image warping. This warping process is identified in literature as `backward warping' where intensity values are interpolated from the input intensity-space and transported to discreet pixel locations in the output image grid. Mathematically, the backward warping process can be written as:

\begin{align}
  & \hat{I}(\textbf{x}) \leftarrow I(\textbf{x} + w(\textbf{x})),\ \forall \textbf{x} \in \Omega_{\hat{I}},
\end{align}

\noindent
where $\Omega_{\hat{I}}$ gives the image domain of output $\hat{I}$, $w(\textbf{x}) = (u,v)$ denotes flow vectors at particular spatial locations $\textbf{x} = (x,y)$ and $\hat{I}$ gives the warped output. Note that $\textbf{x}$ gives pixel locations in the output image grid $\hat{I}$ and the flow is applied (and thus associated) to pixel in $\hat{I}$.

\begin{figure}[h!]
  \centering
    \includegraphics[width=\textwidth]{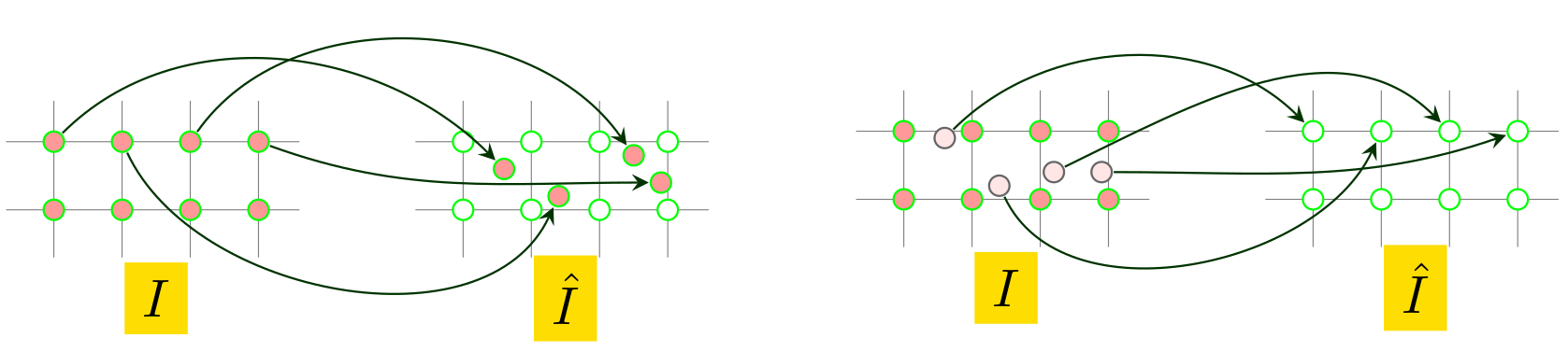}
\caption{\textit{Forward warping} (Left): Some pixels from input $I$ being warped to non-integer output coordinates in I. \textit{Backward Warping} (Right): Some empty pixels in output $\hat{I}$ being populated with values interpolated from non-integer input coordinates in I. Circles represent pixel locations that have a value (filled with colour) or are empty (not filled).}
\label{fig:warping}
\end{figure}

Our proposed layered motion model aims to segment pixels in the input image based on motion. For this, we require a warping process where the flow is associated, and applied, to discreet pixels in the input (not output as in backward warping). Mathematically, the forward warping process is given by:

\begin{align}
  & \hat{I}(\textbf{x} + w(\textbf{x})) \leftarrow I(\textbf{x}),\ \forall \textbf{x} \in \Omega_I,
\end{align}
\noindent
where $\Omega_I$ gives the image domain of input $I$, $w(\textbf{x})$ denotes flow vectors at particular spatial locations $\textbf{x}$ and $\hat{I}$ gives the warped output. Here, the flow is applied to pixels in the input image grid. This enables us to associate flow values to discreet input pixels and thus segregate these input pixels based on their flow. A visual comparison between forward and backward warping is shown in Figure \ref{fig:warping}.

\begin{figure}[t!]
  \centering
    \includegraphics[width=\textwidth]{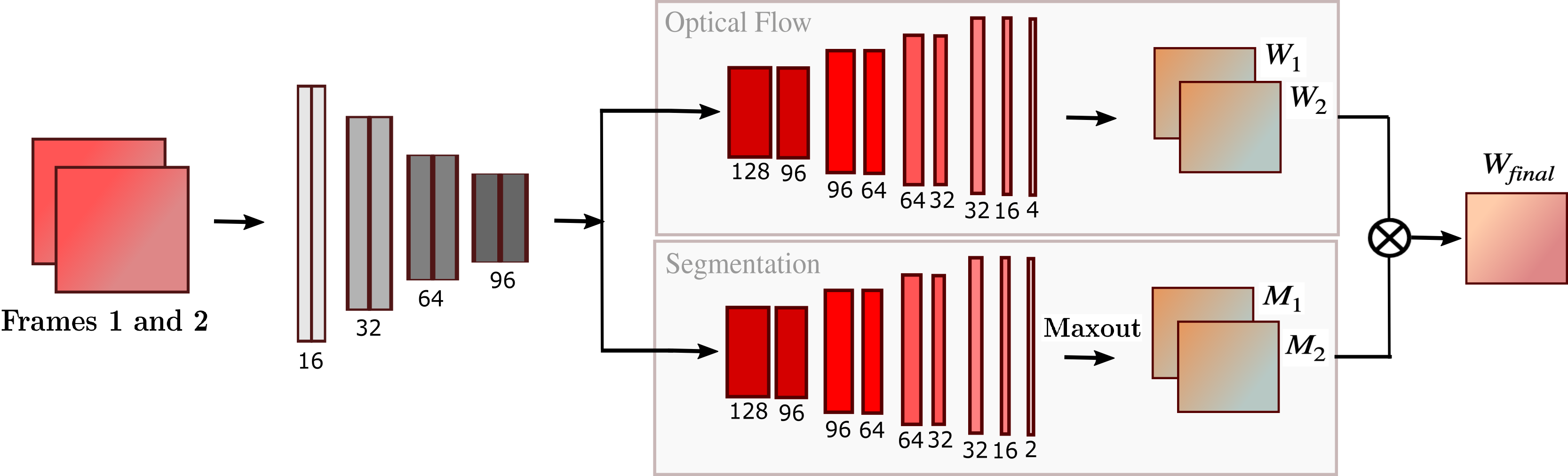}
\caption{Network architecture of maxout baseline incorporating the softmask pipeline from \cite{zhang2018layered}.}
\label{fig:baseline}
\end{figure}

If the flow applied is non-integer valued, warped pixel-values are mapped to locations \textit{between} the pixel grid in the output. To populate the output pixels, each warped pixel-value that ends up between the output grid is interpolated bi-linearly into the surrounding four (empty) output pixels. The final output is obtained after performing summation of pixel-value contributions from all warped input pixels.

Let $V_i(x_i^V, y_i^V)$ represent pixels in the output grid and $W_j(x_j^W, y_j^W)$ represent the warped pixel-values from the input grid. The interpolation process can then be written as:

\begin{align}
	V_i^c = \sum\limits_j^{HW} W_{j}^c & \max(0, 1-\lvert {x_j^W - x_i^V}\rvert)\max(0, 1-\lvert {y_j^W - y_i^V}\rvert) \\ 
	& \forall i \in [1\dots HW],\ \ \forall c \in [R,G,B], \nonumber
\end{align}   
\noindent
where $H$ and $W$ give the height and width of the input and output image grids and $c$ represents the image channels. This sampling process is sub-differentiable and allows gradients to pass through the network facilitating end-to-end learning. It is highly similar in implementation to the sampling mechanism for backward warping outlined in \cite{jaderberg} and can be run very efficiently on GPUs. Our forward warping formulation is equivalent to the ``summation splatting" presented in \cite{niklaus2020softmax}.

\begin{figure*}[t!]
\centering 
	\begin{subfigure}[b]{0.32\linewidth}
		\includegraphics[width=\textwidth]{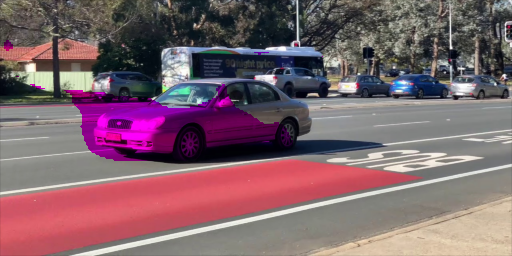}
	\end{subfigure}
	\begin{subfigure}[b]{0.32\linewidth}
		\includegraphics[width=\textwidth]{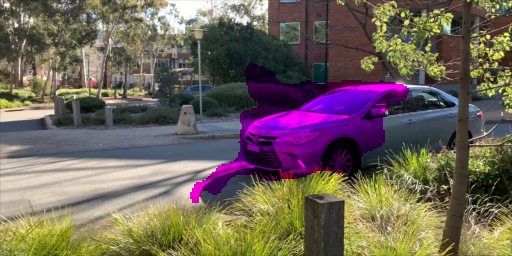}
	\end{subfigure}
	\begin{subfigure}[b]{0.32\linewidth}
		\includegraphics[width=\textwidth]{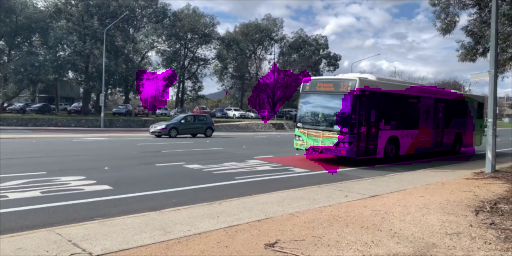}
	\end{subfigure}
	
	\begin{subfigure}[b]{0.32\linewidth}
		\includegraphics[width=\textwidth]{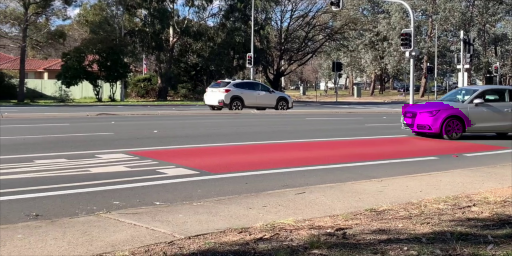}
	\end{subfigure}
	\begin{subfigure}[b]{0.32\linewidth}
		\includegraphics[width=\textwidth]{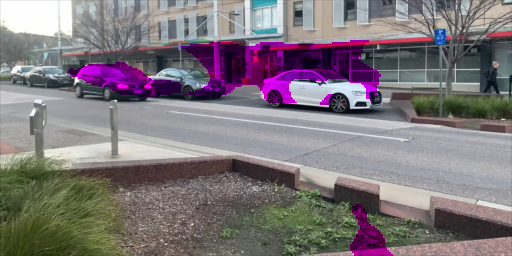}
	\end{subfigure}
	\begin{subfigure}[b]{0.32\linewidth}
		\includegraphics[width=\textwidth]{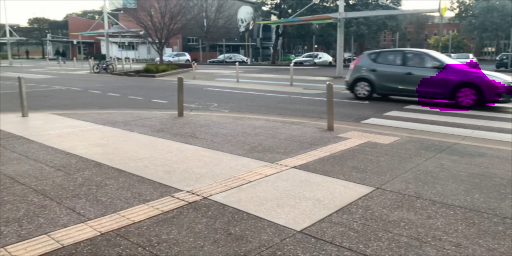}
	\end{subfigure}
	
\caption{Results on the MovingCars dataset of LayerSegNet (ours) showing some failure cases where the dominant foreground moving object is not identified correctly. This usually happens when the object motion cannot be handled by the network's affine parametric model.}


\label{fig:failures-supp}
\end{figure*}

\begin{figure}[t!]
\includegraphics[width=\textwidth]{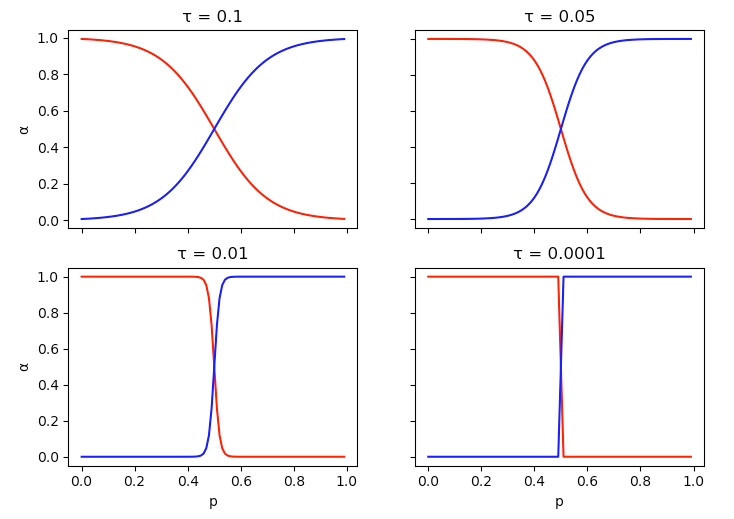}
\caption{Figure showing the output of softmax binning for different $\tau$ values. X-axis gives the input variable p and y-axis gives the $\alpha$ values after softmax for layer-0 (red) and layer-1 (blue).}
\label{fig:softmax_binning}
\end{figure}

\subsection{Maxout Baseline Network Details}

We create a baseline network for motion segmentation following steps to implement the soft-mask pipeline from \cite{zhang2018layered}. We change the optical flow pipeline of our LayerSegNet to output two dense flow maps ($W_1$ and $W_2$) instead of 2 affine parameters. The segmentation module structure remains unchanged: the output is still two segmentation masks. However, we set all activation functions to leaky ReLU (as in optical flow pipeline) including in the last layer i.e. the values of the segmentation masks are not capped to $[0,1]$ as in our LayerSegNet. This is as outlined in \cite{zhang2018layered}.

The segmentation masks are passed through the maxout operation to obtain two disjoint masks $M_1$ and $M_2$. These are multiplied and added to $W_1$ and $W_2$ to obtain the final flow map:
\begin{align}
    W_{final} = W_1*M_1 + W_2*M_2
\end{align}

We pass $W_{final}$ to a supervised loss where it is compared against ground-truth flow. We show the network structure in Figure \ref{fig:baseline}.

\subsection{MovingCars Dataset}
We take videos on a smartphone to capture our MovingCars Dataset by following the trajectory of the moving car. Our affine parametric model provides strong constraint for segmenting pixels into motion layers but it fails to fully segment, or fails completely, in scenes where the dominant motion is too complex for an affine model. We show some example images in  Figure \ref{fig:failures-supp}.

\subsection{Softmax Binning}
Softmax binning removes the need for two separate variables to decide motion layer classification. Instead, there is only one variable that is mapped to the class label depending on which interval it lies in. 
Here, we use two `bins' for our two motion layers to obtain their alpha maps:

Thus, for each pixel $\textbf{x}$, we bin a continuous valued input $p \in [0,1]$ into one of 2 intervals: $[0.0, 0.5]$ or $(0.5, 1.0]$. We then associate a pixel's layer classification depending on which interval it falls under. First, we map $p$ to the respective intervals: 

\begin{align}
    \alpha_{w, b, \tau}^{i}(p) = (w p - b)/\tau
\end{align}
where $\alpha^i$ gives the mapped value for each interval, $i \in \{0, 1, ..., n-1\}$ gives the interval index and $n \in \mathbb{Z}^{+}$ gives the total number of intervals ($n = 2$ for LDIS), $w_i \in \{1, 2, ... , n\}$ is a constant, $b_i \in \{0, \beta_1, \beta_2, ... , \beta_n\}$ are the cut-off points that separate each of the intervals and $\tau > 0$ is a temperature factor. For our two intervals, $\alpha^0$ and $\alpha^1$: $w_i \in \{1, 2\}$ and $b_i \in \{0, 0.5\}$: 

\begin{align}
    \alpha^0(p) &= (w_0 * p - b_0)/\tau = x/\tau \\
    \alpha^1(p) &= (w_1 * p - b_1)/\tau = (2x - 0.5)/\tau  
\end{align}

We then concatenate and pass these through a softmax operation to get the final motion layer label for all pixels $\textbf{x}$: 
\begin{align}
    [\alpha^0, \alpha^1] = \text{Softmax}([\alpha^0, \alpha^1])
    \label{eq:softmaxbinning}
\end{align}
where $\alpha^0$ and $\alpha^1$ give the respective class label values. As $\tau \rightarrow 0$, the alpha maps simulate a one-hot vector 
. We show this behaviour in Figure \ref{fig:softmax_binning}. We plot the input $p$ in x-axis and the alpha map values in y-axis for both the layers (red curve for layer-0 and blue curve for layer-1). We can see that the outputs for both layers start to resemble a one-hot encoding as $\tau \rightarrow 0$ and the y values stay consistent as $p$ gets closer and closer to the cut-off point.

This interval is determined by the cut-off point which, for only two class labels, is 0.5. So, $\alpha^0(p) \rightarrow 1$ if $\textbf{x}$ belongs to layer-0 and $\alpha^1(p) \rightarrow 1$ if it belongs to layer-1. This binning process also ensures that each pixel $\textbf{x}$ can only belong to a single layer. More details on softmax binning can be found in \cite{yang2018deep}.


\bibliographystyle{splncs}
\bibliography{egbib}

\begin{thebibliography}{10}

\bibitem{layered1994}
Wang, J.Y., Adelson, E.H.:
\newblock Representing moving images with layers.
\newblock IEEE Transactions on Image Processing \textbf{3} (1994)  625--638

\bibitem{koffka2013principles}
Koffka, K.:
\newblock Principles of Gestalt psychology. Volume~44.
\newblock Routledge (2013)

\bibitem{lai2020mast}
Lai, Z., Lu, E., Xie, W.:
\newblock Mast: A memory-augmented self-supervised tracker.
\newblock In: Proc of the IEEE/CVF Conf on CVPR. (2020)  6479--6488

\bibitem{meinhardt2020make}
Meinhardt, T., Leal-Taix{\'e}, L.:
\newblock Make one-shot video object segmentation efficient again.
\newblock arXiv preprint arXiv:2012.01866 (2020)

\bibitem{perazzi2017learning}
Perazzi, F., Khoreva, A., Benenson, R., Schiele, B., Sorkine-Hornung, A.:
\newblock Learning video object segmentation from static images.
\newblock In: Proc of the IEEE conf on CVPR. (2017)  2663--2672

\bibitem{maninis2018video}
Maninis, K.K., Caelles, S., Chen, Y., Pont-Tuset, J., Leal-Taix{\'e}, L.,
  Cremers, D., Van~Gool, L.:
\newblock Video object segmentation without temporal information.
\newblock IEEE tran on pattern analysis and machine intelligence \textbf{41}
  (2018)  1515--1530

\bibitem{mahadevan2020making}
Mahadevan, S., Athar, A., O{\v{s}}ep, A., Hennen, S., Leal-Taix{\'e}, L.,
  Leibe, B.:
\newblock Making a case for 3d convolutions for object segmentation in videos.
\newblock arXiv preprint arXiv:2008.11516 (2020)

\bibitem{yang2019anchor}
Yang, Z., Wang, Q., Bertinetto, L., Hu, W., Bai, S., Torr, P.H.:
\newblock Anchor diffusion for unsupervised video object segmentation.
\newblock In: Proc of the IEEE/CVF Int Conf on Computer Vision. (2019)
  931--940

\bibitem{ren2021reciprocal}
Ren, S., Liu, W., Liu, Y., Chen, H., Han, G., He, S.:
\newblock Reciprocal transformations for unsupervised video object
  segmentation.
\newblock In: Proc of the IEEE/CVF Conf on CVPR. (2021)  15455--15464

\bibitem{lu2020learning}
Lu, X., Wang, W., Shen, J., Tai, Y.W., Crandall, D.J., Hoi, S.C.:
\newblock Learning video object segmentation from unlabeled videos.
\newblock In: Proc of the IEEE/CVF conf on CVPR. (2020)  8960--8970

\bibitem{darrell1991robust}
Darrell, T., Pentland, A.:
\newblock Robust estimation of a multi-layered motion representation.
\newblock In: Proc of the IEEE Workshop on Visual Motion, IEEE (1991)  173--178

\bibitem{jepson1993mixture}
Jepson, A., Black, M.J.:
\newblock Mixture models for optical flow computation.
\newblock In: Proc of IEEE Conf on CVPR, IEEE (1993)  760--761

\bibitem{sevilla}
Sevilla-Lara, L., Sun, D., Jampani, V., Black, M.J.:
\newblock Optical flow with semantic segmentation and localized layers.
\newblock In: Proc of the IEEE Conf on CVPR. (2016)  3889--3898

\bibitem{zhang2018layered}
Zhang, X., Ma, D., Ouyang, X., Jiang, S., Gan, L., Agam, G.:
\newblock Layered optical flow estimation using a deep neural network with a
  soft mask.
\newblock In: Proc of the Twenty-Seventh Int Joint Conf on Artificial
  Intelligence, {IJCAI-18}, Int Joint Conferences on Artificial Intelligence
  Organization (2018)  1170--1176

\bibitem{yang2021self}
Yang, C., Lamdouar, H., Lu, E., Zisserman, A., Xie, W.:
\newblock Self-supervised video object segmentation by motion grouping.
\newblock arXiv preprint arXiv:2104.07658 (2021)

\bibitem{wang2015saliency}
Wang, W., Shen, J., Porikli, F.:
\newblock Saliency-aware geodesic video object segmentation.
\newblock In: Proc of the IEEE conf on CVPR. (2015)  3395--3402

\bibitem{guo2008spatio}
Guo, C., Ma, Q., Zhang, L.:
\newblock Spatio-temporal saliency detection using phase spectrum of quaternion
  fourier transform.
\newblock In: 2008 IEEE Conf on CVPR, IEEE (2008)  1--8

\bibitem{mahadevan2009spatiotemporal}
Mahadevan, V., Vasconcelos, N.:
\newblock Spatiotemporal saliency in dynamic scenes.
\newblock IEEE tran on pattern analysis and machine intelligence \textbf{32}
  (2009)  171--177

\bibitem{lee2011key}
Lee, Y.J., Kim, J., Grauman, K.:
\newblock Key-segments for video object segmentation.
\newblock In: 2011 Int conf on computer vision, IEEE (2011)  1995--2002

\bibitem{ma2012maximum}
Ma, T., Latecki, L.J.:
\newblock Maximum weight cliques with mutex constraints for video object
  segmentation.
\newblock In: 2012 IEEE Conf on CVPR, IEEE (2012)  670--677

\bibitem{zhang2013video}
Zhang, D., Javed, O., Shah, M.:
\newblock Video object segmentation through spatially accurate and temporally
  dense extraction of primary object regions.
\newblock In: Proc of the IEEE conf on CVPR. (2013)  628--635

\bibitem{brox2010object}
Brox, T., Malik, J.:
\newblock Object segmentation by long term analysis of point trajectories.
\newblock In: European conference on computer vision, Springer (2010)  282--295

\bibitem{ochs2011object}
Ochs, P., Brox, T.:
\newblock Object segmentation in video: a hierarchical variational approach for
  turning point trajectories into dense regions.
\newblock In: 2011 Int Conf on Computer Vision, IEEE (2011)  1583--1590

\bibitem{li2018video}
Li, X., Loy, C.C.:
\newblock Video object segmentation with joint re-identification and
  attention-aware mask propagation.
\newblock In: Proc of the ECCV. (2018)  90--105

\bibitem{oh2018fast}
Oh, S.W., Lee, J.Y., Sunkavalli, K., Kim, S.J.:
\newblock Fast video object segmentation by reference-guided mask propagation.
\newblock In: Proc of the IEEE Conf on CVPR. (2018)  7376--7385

\bibitem{tokmakov2019learning}
Tokmakov, P., Schmid, C., Alahari, K.:
\newblock Learning to segment moving objects.
\newblock Int Journal of Computer Vision \textbf{127} (2019)  282--301

\bibitem{jain2017fusionseg}
Jain, S.D., Xiong, B., Grauman, K.:
\newblock Fusionseg: Learning to combine motion and appearance for fully
  automatic segmentation of generic objects in videos.
\newblock In: 2017 IEEE conf on CVPR, IEEE (2017)  2117--2126

\bibitem{zhou2020motion}
Zhou, T., Wang, S., Zhou, Y., Yao, Y., Li, J., Shao, L.:
\newblock Motion-attentive transition for zero-shot video object segmentation.
\newblock In: Proc of the AAAI Conf on Artificial Intelligence. Volume~34.
  (2020)  13066--13073

\bibitem{fragkiadaki2015learning}
Fragkiadaki, K., Arbelaez, P., Felsen, P., Malik, J.:
\newblock Learning to segment moving objects in videos.
\newblock In: Proc of the IEEE Conf on CVPR. (2015)  4083--4090

\bibitem{lu2019see}
Lu, X., Wang, W., Ma, C., Shen, J., Shao, L., Porikli, F.:
\newblock See more, know more: Unsupervised video object segmentation with
  co-attention siamese networks.
\newblock In: Proc of the IEEE/CVF Conf on CVPR. (2019)  3623--3632

\bibitem{locatello2020object}
Locatello, F., Weissenborn, D., Unterthiner, T., Mahendran, A., Heigold, G.,
  Uszkoreit, J., Dosovitskiy, A., Kipf, T.:
\newblock Object-centric learning with slot attention.
\newblock arXiv preprint arXiv:2006.15055 (2020)

\bibitem{ayer1995layered}
Ayer, S., Sawhney, H.S.:
\newblock Layered representation of motion video using robust
  maximum-likelihood estimation of mixture models and mdl encoding.
\newblock In: Proc of IEEE Int Conf on Computer Vision, IEEE (1995)  777--784

\bibitem{hsu1994accurate}
Hsu, S., Anandan, P., Peleg, S.:
\newblock Accurate computation of optical flow by using layered motion
  representations.
\newblock In: Proc of 12th Int Conf on Pattern Recognition. Volume~1., IEEE
  (1994)  743--746

\bibitem{sun2013fully}
Sun, D., Wulff, J., Sudderth, E.B., Pfister, H., Black, M.J.:
\newblock A fully-connected layered model of foreground and background flow.
\newblock In: Proc of the IEEE Conf on CVPR. (2013)  2451--2458

\bibitem{lee1997layered}
Lee, M.C., Chen, W.G., Lin, C.l.B., Gu, C., Markoc, T., Zabinsky, S.I.,
  Szeliski, R.:
\newblock A layered video object coding system using sprite and affine motion
  model.
\newblock IEEE Transactions on circuits and systems for video technology
  \textbf{7} (1997)  130--145

\bibitem{arrigoni2019robust}
Arrigoni, F., Pajdla, T.:
\newblock Robust motion segmentation from pairwise matches.
\newblock In: Proc of the IEEE Int Conf on Computer Vision. (2019)  671--681

\bibitem{jung2014rigid}
Jung, H., Ju, J., Kim, J.:
\newblock Rigid motion segmentation using randomized voting.
\newblock In: Proc of the IEEE Conf on CVPR. (2014)  1210--1217

\bibitem{ji2015shape}
Ji, P., Salzmann, M., Li, H.:
\newblock Shape interaction matrix revisited and robustified: Efficient
  subspace clustering with corrupted and incomplete data.
\newblock In: Proc of the IEEE Int Conf on computer Vision. (2015)  4687--4695

\bibitem{fischler1981random}
Fischler, M.A., Bolles, R.C.:
\newblock Random sample consensus: a paradigm for model fitting with
  applications to image analysis and automated cartography.
\newblock Communications of the ACM \textbf{24} (1981)  381--395

\bibitem{verri1989motion}
Verri, A., Uras, S., De~Micheli, E.:
\newblock Motion segmentation from optical flow.
\newblock In: Alvey Vision Conference. (1989)  1--6

\bibitem{zhuo2019unsupervised}
Zhuo, T., Cheng, Z., Zhang, P., Wong, Y., Kankanhalli, M.:
\newblock Unsupervised online video object segmentation with motion property
  understanding.
\newblock IEEE Transactions on Image Processing \textbf{29} (2019)  237--249

\bibitem{ranjan2019competitive}
Ranjan, A., Jampani, V., Balles, L., Kim, K., Sun, D., Wulff, J., Black, M.J.:
\newblock Competitive collaboration: Joint unsupervised learning of depth,
  camera motion, optical flow and motion segmentation.
\newblock In: Proc of the IEEE Conf on CVPR. (2019)  12240--12249

\bibitem{yang2018deep}
Yang, Y., Morillo, I.G., Hospedales, T.M.:
\newblock Deep neural decision trees.
\newblock arXiv preprint arXiv:1806.06988 (2018)

\bibitem{niklaus2020softmax}
Niklaus, S., Liu, F.:
\newblock Softmax splatting for video frame interpolation.
\newblock In: Proc of the IEEE/CVF Conf on CVPR. (2020)  5437--5446

\bibitem{bruhn2005lucas}
Bruhn, A., Weickert, J., Schn{\"o}rr, C.:
\newblock Lucas/kanade meets horn/schunck: Combining local and global optic
  flow methods.
\newblock Int journal of computer vision \textbf{61} (2005)  211--231

\bibitem{dosovitskiy2015flownet}
Dosovitskiy, A., Fischer, P., Ilg, E., Hausser, P., Hazirbas, C., Golkov, V.,
  van~der Smagt, P., Cremers, D., Brox, T.:
\newblock Flownet: Learning optical flow with convolutional networks.
\newblock In: Proc of the IEEE Int Conf on Computer Vision. (2015)  2758--2766

\bibitem{pwcnet}
Sun, D., Yang, X., Liu, M.Y., Kautz, J.:
\newblock Pwc-net: Cnns for optical flow using pyramid, warping, and cost
  volume.
\newblock In: Proc of the IEEE Conf on CVPR. (2018)  8934--8943

\bibitem{maas2013rectifier}
Maas, A.L., Hannun, A.Y., Ng, A.Y.:
\newblock Rectifier nonlinearities improve neural network acoustic models.
\newblock In: Proc. icml. Volume~30. (2013) ~3

\bibitem{nair2010rectified}
Nair, V., Hinton, G.E.:
\newblock Rectified linear units improve restricted boltzmann machines.
\newblock In: Proc of the 27th Int Conf on machine learning (ICML-10). (2010)
  807--814

\bibitem{godin2018dual}
Godin, F., Degrave, J., Dambre, J., De~Neve, W.:
\newblock Dual rectified linear units (drelus): a replacement for tanh
  activation functions in quasi-recurrent neural networks.
\newblock Pattern Recognition Letters \textbf{116} (2018)  8--14

\bibitem{janai}
Janai, J., Fatma, G., Black, M., Geiger, A.:
\newblock Unsupervised learning of multi-frame optical flow with occlusions.
\newblock In: ECCV. (2018)

\bibitem{kingma2014adam}
Kingma, D.P., Ba, J.:
\newblock Adam: A method for stochastic optimization.
\newblock arXiv preprint arXiv:1412.6980 (2014)

\bibitem{jaderberg}
Jaderberg, M., Simonyan, K., Zisserman, A.,  et~al.:
\newblock Spatial transformer networks.
\newblock In: Advances in neural information processing systems. (2015)
  2017--2025

\end{thebibliography}

\end{document}